\documentclass[10pt,twocolumn,letterpaper]{article}
\pdfoutput=1
\usepackage{iccv}
\usepackage{times}
\usepackage{epsfig}
\usepackage{graphicx}
\usepackage{amsmath}
\usepackage{amssymb}

\usepackage{booktabs}
\usepackage{bbm}
\usepackage{multirow}
\usepackage{makecell}
\usepackage{float}
\restylefloat{table}
\usepackage{caption}
\usepackage{subcaption}
\usepackage[normalem]{ulem}

\usepackage[pagebackref=true,breaklinks=true,letterpaper=true,colorlinks,bookmarks=false]{hyperref}
\usepackage{pifont}
\iccvfinalcopy 

\newcommand{\softmax}{\operatorname{softmax}}
\newcommand{\TopK}{\operatorname{TopK}}

\newcommand{\MLP}{\operatorname{MLP}}

\def\iccvPaperID{5169} 

\begin{document}

\title{Enhancing NeRF akin to Enhancing LLMs: Generalizable NeRF Transformer with Mixture-of-View-Experts}

\author{Wenyan Cong$^{1}$\thanks{Equal contribution.}, Hanxue Liang$^{2,1}$\footnotemark[1], Peihao Wang$^{1}$, Zhiwen Fan$^{1}$, Tianlong Chen$^{1}$,\\
Mukund Varma$^{3,1}$, Yi Wang$^{1}$, Zhangyang Wang$^{1}$\\
$^{1}$University of Texas at Austin, 
$^{2}$University of Cambridge, 
$^{3}$Indian Institute of Technology Madras\\
\tt\small \{wycong, peihaowang, zhiwenfan, tianlong.chen, panzer.wy, atlaswang\}@utexas.edu,\\
\tt\small hl589@cam.ac.uk, mukundvarmat@gmail.com
}

\maketitle


\begin{abstract}
  Cross-scene generalizable NeRF models, which can directly synthesize novel views of unseen scenes, have become a new spotlight of the NeRF field. Several existing attempts rely on increasingly end-to-end ``neuralized'' architectures, i.e., replacing scene representation and/or rendering modules with performant neural networks such as transformers, and turning novel view synthesis into a feed-forward inference pipeline. While those feedforward ``neuralized'' architectures still do not fit diverse scenes well out of the box, we propose to bridge them with the powerful Mixture-of-Experts (MoE) idea from large language models (LLMs), which has demonstrated superior generalization ability by balancing between larger overall model capacity and flexible per-instance specialization. Starting from a recent generalizable NeRF architecture called GNT \cite{gnt}, we first demonstrate that MoE can be neatly plugged in to enhance the model. We further customize a shared permanent expert and a geometry-aware consistency loss to enforce cross-scene consistency and spatial smoothness respectively, which are essential for generalizable view synthesis. Our proposed model, dubbed GNT with Mixture-of-View-Experts (\textbf{GNT-MOVE}), has experimentally shown state-of-the-art results when transferring to unseen scenes, indicating remarkably better cross-scene generalization in both zero-shot and few-shot settings. Our codes are available at \url{https://github.com/VITA-Group/GNT-MOVE}.

\end{abstract}


\section{Introduction}
\label{sec:intro}

Given several images from different viewpoints, Neural Radiance Field (NeRF) has achieved remarkable success on synthesizing novel views. Most existing methods \cite{NERF, liu2020neural, reiser2021kilonerf, NeX,plenoxels, directvoxgo, barron2021mip,xu2022sinnerf,xu2023neurallift,jiang2023alignerf} focus on overfitting one single scene by reconstructing its 3D radiance field in a ``backward'' manner. Though capable of generating realistic and consistent novel views, the need for retraining on each new scene limits their practical applications. Recently, generalizable NeRF has settled a new trend: in place of the costly per-scene fitting, several pioneer works \cite{yu2021pixelnerf, wang2021ibrnet, yao2018mvsnet, gnt, gpnr} attempt to synthesize novel views of unseen scenes in a ``feedforward'' fashion on the fly.
Those models are first pre-trained by learning how to represent scenes and render novel views from captured images across different scenes, achieving high-quality ``zero-shot" inference results on new scenes.
Among them, Generalizable NeRF Transformer (GNT)~\cite{gnt} stands out by replacing the explicit scene modeling and rendering function via unified, data-driven, and scalable transformers, and automatically inducing multi-view consistent geometries and renderings via large-scale novel view synthesis pre-training.

However, those cross-scene NeRF models face the fundamental dilemma between \textbf{``generality''} and \textbf{``specialization''}. On the one hand, they need to broadly cover both diverse scene representations and/or rendering mechanisms due to different scene properties (\emph{e.g.}, color, materials) 
-- hence larger overall model size is needed to guarantee sufficient expressiveness. On the other hand, since a single scene usually consists of specialized self-similar appearance patterns, those models must also be capable of per-scene specialization to model the scene closely. Existing generalizable models still do not achieve a satisfactory balance between both ``generality'' and ``specialization'', as most of them~\cite{gnt, gpnr} do not fit diverse scenes well out of box, and some~\cite{wang2021ibrnet, yao2018mvsnet} will need extra per-scene optimization step.

To fill the aforementioned gap, we propose to introduce and customize the powerful Mixture-of-Experts (MoE) idea \cite{shazeer2017outrageously} into GNT framework, which is composed of a \textit{view transformer} that aggregates multi-view image features and a \textit{ray transformer} that decodes the point feature to synthesize novel views.
The inspiration is drawn from \textit{Large Language Models} (\textbf{LLMs}) ~\cite{lepikhin2020gshard,fedus2021switch}, where MoE has become the key knob to improve the generalization of these models, scaling up the total model size without exploding the per-inference cost, by encouraging different submodels (combination of activated experts) to be sparsely activated for different inputs and hence become ``specialized".

Specifically, to balance between cross-scene ``generalization" and per-scene ``specialization", we bake MoE into GNT's view transformer\footnote{In this paper, we mainly focus on the view transformer based on the hypothesis that the modular design of MoE could be naturally beneficial to multi-view feature aggregation. Introducing MoE into the ray transformer may be also promising and we leave it as future work.}, leading to a new GNT with Mixture-of-View-Experts (\textbf{GNT-MOVE}). However, as we observed from experiments, naively plugging MoE into NeRF fails to well balance between generality and specialization, due to their intension with generalizable NeRF's cross-scene \textbf{consistency} and spatial \textbf{smoothness} priors:
\begin{itemize} \vspace{-1.5mm}
    \item Cross-scene \textbf{consistency}: similar appearance patterns or similar materials, \underline{from different scenes}, should be treated consistently by choosing similar experts. \vspace{-1mm} 
    \item Spatial \textbf{smoothness}: nearby views \underline{in the same scene} should change continuously \& smoothly, hereby making similar or smoothly transiting expert selection.\vspace{-1mm} 
\end{itemize}
Those two ``priors" are owing to the natural image rendering and multi-view geometry constraints. Yet, enforcing them risks causing the notorious representation collapse of MoEs \cite{zhou2022mixture}, i.e., differently activated submodels may naively learn the same or similar functions and be unable to capture diverse specialized features. Such representational collapse has been addressed a lot in the general MoE literature \cite{zuo2021taming,lewis2021base,roller2021hash}. But it remains elusive whether those solutions will be at odds with the ``consistency/smoothness": \textit{a new challenge we must pay attention to}.

In order to mitigate such gaps, we investigate two customized improvements of MoE for NeRF. \underline{Firstly}, we augment the MoE layer with a shared permanent expert, that will be selected in all cases. This shared expert enforces the commodity across scenes as an architectural regularization, and boosts cross-scene consistency. \underline{Secondly}, a spatial smoothness objective is introduced for geometric-aware continuity, by encouraging two spatially close points to choose similar experts, and using the geometric distance between sampled points to re-weight their expert selections. We empirically find the two consistency regularizations to work well with the typical expert diversity regularizer in MoEs, together ensuring effectively large model capacity as well as meeting the consistency/smoothness demands. We have conducted comprehensive experiments on complex scene benchmarks. Remarkably, when trained on multiple scenes, GNT-MOVE attains state-of-the-art performance in two aspects: (1) often notably better zero-shot generalization to unseen scenes; and (2) consistently stronger performance on few-shot generalization to unseen scenes.

Our main contributions can be summarized as follows: \vspace{-1.5mm}
\begin{itemize}
\item 
We present an LLM-inspired NeRF framework, GNT-MOVE, which significantly pushes the frontier of generalizable novel view synthesis on complex scenes by introducing Mixture-of-Experts (MoE) transformers. \vspace{-1mm}

\item 
To tailor MoE for generalizable NeRF, we introduce a shared permanent expert for cross-scene rendering consistency, and a geometry-aware spatial consistency objective for cross-view spatial smoothness. \vspace{-1mm}   

\item Experiments on complex scene benchmarks validate the effectiveness of GNT-MOVE on cross-scene generalization with both zero-shot and few-shot settings.\vspace{-1mm}

\end{itemize}

\begin{figure*}[tp]
    \begin{centering}
        \vspace{-1.5mm}
        \includegraphics[width=1.0\linewidth]{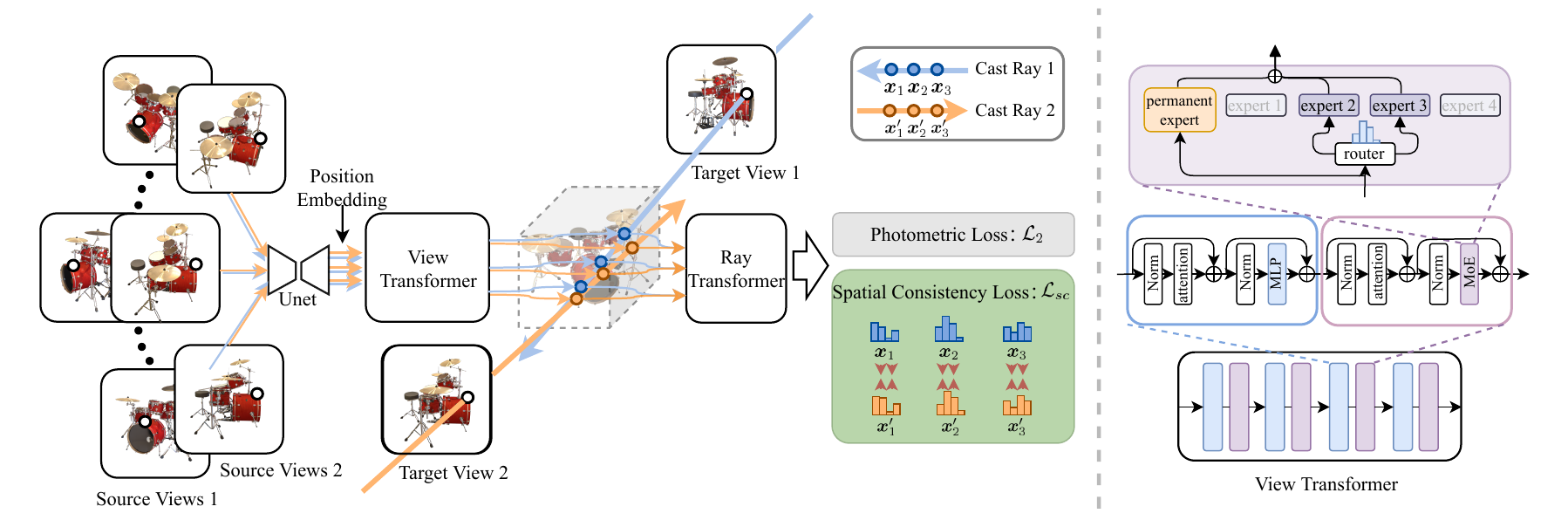}
        \par
    \end{centering}
    \vspace{-1.5mm}
    \caption{\textbf{Overview of our GNT-MOVE.} \textit{Left sub-figure}: for each ray in the target view, sampled points will aggregate multi-view features from source views by passing through the view transformer. \textit{Right sub-figure}: in view transformer, we embed the MoE layer in the transformer blocks. Point token will be processed by both router-selected experts and our proposed permanent expert to enforce cross-scene consistency. Note that we use 4 MoE embedded transformer blocks, and 4 experts per MoE layer, leading to ${{4}\choose{2}}^4$ = 1,296 total expert combinations to provide sufficiently large and diverse coverage. 
    }
    \vspace{-2mm}
    \label{fig:framework}
\end{figure*}

\section{Related Works}

\paragraph{NeRF and Its Generalization.} Novel View Synthesis (NVS) aims to generate unseen views given a set of posed images. 
Recently, Neural Radiance Field (\emph{i.e.}, NeRF~\cite{NERF}) has achieved remarkable performance on novel view synthesis by volume rendering on a radiance field. Several followups extend NeRF by proposing new parameterizations of rays~\cite{barron2021mip, barron2022mip} to improve rendering quality, using explicit data structures or distillation~\cite{ fridovich2022plenoxels,liu2020neural,sun2022direct,muller2022instant,wang2022r2l, garbin2021fastnerf,hedman2021baking} to improve efficiency, or adopting spatial-temporal modeling~\cite{park2021nerfies,pumarola2021d,gafni2021dynamic,li2021neural,xian2021space} to extend it to dynamic scenarios.

However, the original NeRF needs to retrain on each new scene, thus limiting its practical applications. To tackle the cross-scene generalization, one line of works~\cite{trevithick2021grf, jang2021codenerf, yu2021pixelnerf} incorporate a convolutional encoder and use the same MLP conditioned on different image features to model different objects. More recently, another line of works~\cite{gnt,suhail2022generalizable,kulhanek2022viewformer,paul2021transnerf,zhao2022end,wang2021ibrnet} adopt transformer-based network with epipolar constraints to synthesize novel views of unseen scenes in a "feedforward" fashion on the fly. Our method is also based on the transformer to render novel scenes in a feedforward fashion. The difference is that we customized the powerful MoE ideas into our framework to balance between cross-scene generalization and per-scene specialization, thus capable of modeling diverse complex scenes and rendering more realistic results, in few- or zero-shot.

\vspace{-0.7em}
\paragraph{Mixture-of-Experts (MoE).}
MoEs~\cite{jacobs1991adaptive,jordan1994hierarchical,chen1999improved,6215056,roller2021hash,fan2022m3vit,chen2022sparse} perform input-dependent computations with a combination of sub-models (a.k.a. experts) according to certain learned or ad-hoc routing policies~\cite{dua2021tricks,roller2021hash}. Various successful cases of MoE have been shown in a wide range of applications. Recent advances~\cite{shazeer2017outrageously,lepikhin2020gshard,fedus2021switch,shen2023flan} in the natural language processing field propose sparse-gated MoEs to scale up LLM capacity without sacrificing per-inference cost and encourage different modules with distinct functionalities. This helps to unleash the massive potential for compositional unseen generalization \cite{yang2022deep,liu2022dataset,yang2022KF} besides excellent accuracy-efficiency trade-offs. MoE also gains popularity in computer vision~\cite{abbas2020biased,fan2022m3vit,pavlitskaya2020using}, although most works~\cite{eigen2013learning,ahmed2016network,gross2017hard,wang2020deep,yang2019condconv} only focus on classification tasks. 

A few works have explored the sparsely activated sub-models idea implicitly in NeRF. Kilo-NeRF~\cite{reiser2021kilonerf} introduces thousands of tiny MLPs 
to divide and conquer the entire scene modeling. Block-NeRF~\cite{tancik2022block} enables NeRF to represent a street-scale scene by dividing large environments into individually trained NeRFs. NID~\cite{wang2022neural} improves both data and training efficiency of INR by assembling a group of coordinate-based sub-networks. NeurMiPs~\cite{lin2022neurmips} leverages a collection of local planar experts in 3D space to boost the reconstruction quality. Different from all previous arts trying to improve per-scene rendering or fitting, we make the first attempt to customize MoE for generalizable NeRF and improve its performance on rendering novel unseen scenes. 

\section{Preliminary}

\paragraph{GNT}
Generalizable NeRF Transformer (GNT)~\cite{gnt} is a pure, unified transformer-based architecture that efficiently reconstructs Neural Radiance Fields (NeRFs) on the fly from source views. It is composed of two transformer-based stages. In the first stage, the \textit{view transformer} predicts coordinate-aligned features for each point by aggregating information from epipolar lines of its neighboring views. In the second stage, the \textit{ray transformer} composes point-wise features along the ray to compute the ray color. More precisely, given $N$ source images 
$\{\mathbf{I}_i \in\mathbbm{R}^{H\times W\times 3}\}_{i=1}^{N}$, for each sampled point $\boldsymbol{x} \in\mathbbm{R}^{3}$ on a ray emitted from the target view, the view transformer is formulated as:
\vspace{-1mm}
\begin{equation}
\begin{split}
    \mathcal{F}(\boldsymbol{x},\boldsymbol{\theta}) =\text{V-Trans}(\mathbf{F}(\Pi_{1}(\boldsymbol{x})), ...,\mathbf{F}(\Pi_{N}(\boldsymbol{x}))),
\end{split}
\end{equation}
where $\Pi_{i}(\boldsymbol{x})$ is to project 3D point $\boldsymbol{x}$ onto the $i$-th image plane $\mathbf{I}_{i}$, and $\mathbf{F}$ is a small U-Net~\cite{ronneberger2015u} based CNN that interpolates features at the projected image point. The view transformer is adopted to combine all the extracted features into a coordinate-aligned feature volume.

These multi-view aggregated features are then fed into the ray transformer. The ray transformer then performs mean pooling over the predicted tokens and map them to RGB via an MLP to obtain the rendered ray color:
\vspace{-1mm}
\begin{equation}
    \mathbf{\mathcal{C}}(r) =  \MLP \circ\;\text{R-Trans}(\mathcal{F}(\boldsymbol{x}_1, \boldsymbol{\theta}),..., \mathcal{F}(\boldsymbol{x}_{M}, \boldsymbol{\theta})).
\end{equation}
$\left\{\boldsymbol{x}_{1},...,\boldsymbol{x}_{M}\right\}$ are 3D points sampled along the same ray $r$. In this work, we choose GNT as the backbone due to its outstanding performance. However, our methodology shall be general to other transformer-based NeRFs~\cite{suhail2022generalizable,kulhanek2022viewformer,paul2021transnerf}
\vspace{-2mm}

\paragraph{MoE}
A Mixture of Experts (MoE) layer typically contains a group of $E$ experts $f_1,f_2,\cdots,f_{E}$ and a router $\mathcal{R}$ whose output is an $E$-dimensional vector. The expert networks are in the form of a multi-layer perception~\cite{fedus2021switch,riquelme2021scaling} in ViTs.  
The router $\mathcal{R}$ plays the role of expert selection, and we adopt a representative router called top-$K$ gating~\cite{shazeer2017outrageously}. With input token $\mathbf{x}$, the
resultant output $\mathbf{y}$ of MoE layers can be formulated as the summation of the selected top $K$ experts
from $E$ expert candidates using a router:
\vspace{-2mm}
\begin{equation}
\begin{split}
\mathbf{y}&=\sum_{e=1}^{E}\mathcal{R}(\mathbf{x})_e\cdot f_e(\mathbf{x}), \\
\mathcal{R}(\mathbf{x})&=\softmax(\TopK(\mathcal{G}(\mathbf{x}),K)), \\
\TopK(\mathbf{v},K)_i&= \left\{\begin{array}{ll}
v_i & \text{if $v_i$ is in the top $K$ elements of $\mathbf{v}$} \\
0 & \text{otherwise}
\end{array}\right.
\end{split}
\label{eqn:moe_output}
\end{equation}
where $\mathcal{G}$ represents the learnable network within the router.


\section{Method}

\paragraph{Overview.}
We scale up GNT model with MoE layer in this section. The main pipeline is illustrated in Figure \ref{fig:framework}.
Our design principle is that we only make necessary and minimal modifications to the vanilla GNT to preserve its standardized architecture and ease of use.

\begin{figure}[t]
    \centering
    \includegraphics[width=1.0\linewidth]{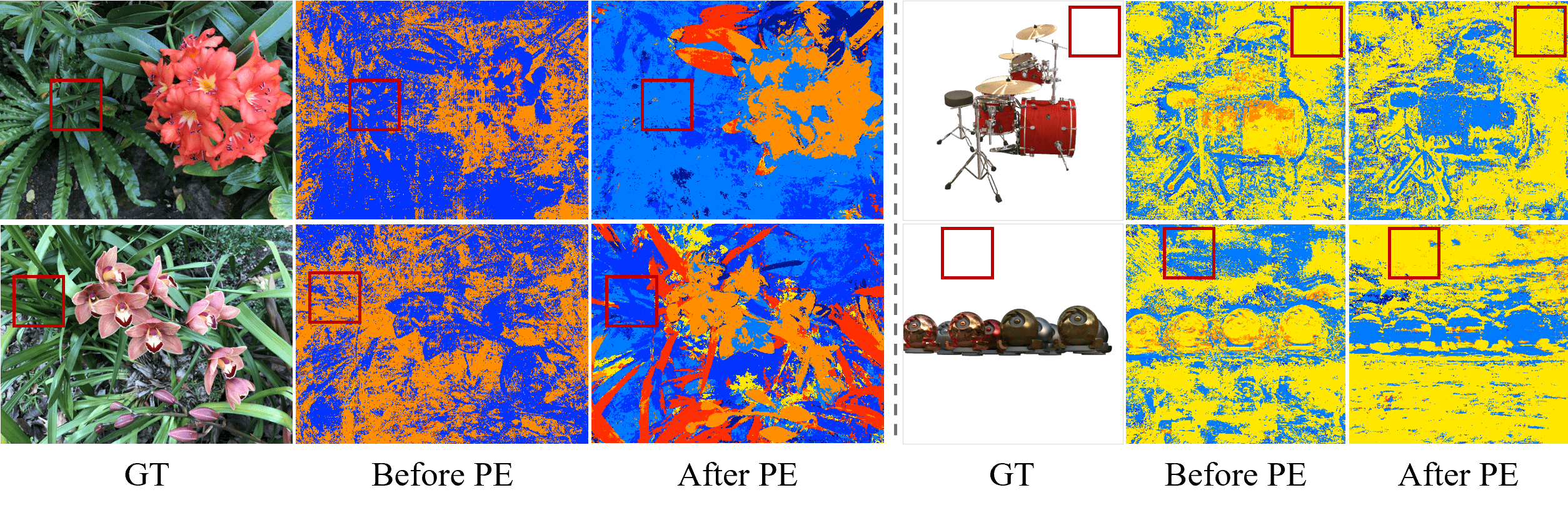}
    \caption{\textbf{Cross-scene inconsistency.} For similar colors or patterns from different scenes (left: green leaves in Flower and Orchids, right: white background in Drums and Materials), the router selects different experts (visualized with different colors). 
    A permanent expert enforces commodity across scenes to enhance cross-scene consistency.
    }
    \vspace{-2mm}
    \label{fig:permemant_expert}
\end{figure}

\subsection{Mixture of View Experts: The Basic Pipeline} \label{sec:pipeline}

It is discussed in \cite{gnt} that GNT leverages the UNet to extract geometry, appearance, and local light transport information from the 2D images, and view transformer will integrate those features to estimate the point-wise rendering parameters (such as occupancy, transparency, and reflectance) on the latent space for the ray transformer.
We notice that the natural shading properties are often exclusive to each other and thus sparsely activated (e.g., diffuse reflection vs. specular reflection).
Also in typical rendering engines, displaying a scene usually invokes different graphical shaders to handle spatially varying materials.
These observations altogether motivate us to plug MoE modules into the view transformer to specialize different components for specific rendering properties.

Our pipeline could be seen in Figure \ref{fig:framework}. As shown in the right sub-figure, in the view transformer, we replace the dense MLP layer with a sparsely activated MoE layer composed of a set of half-sized MLP experts $\{f_e\}_{e=1}^{E}$.
As in Equation \ref{eqn:moe_output}, the output of each MoE layer is the weighted summation of the outputs from the selected top $K$ experts.
Considering the view transformer with $L$ MoE-embedded transformer blocks, we note that the number of possible expert combinations can factually reach ${E \choose K}^L$, which can provide sufficiently broad and diverse coverage.

Following many MoE prior arts \cite{zuo2021taming,lewis2021base,roller2021hash}, we also enforce balanced and diverse expert usage to avoid representation collapse. 
Particularly, within each training batch, we sample 3D points $\boldsymbol{x}$ from ray group $\mathbf{R}$, where the rays are emitted from multiple different views of the same scene, and regularize expert selection via Coefficient of Variation (CV) of the sparse routing~\cite{shazeer2017outrageously}:
\vspace{-1mm}
\begin{equation}
\begin{split}
     \mathcal{L}_{div}&=CV(\mathop{\mathbb{E}}_{r\sim \mathbf{R}} \mathop{\mathbb{E}}_{\boldsymbol{x}\in r} \mathcal{R}(\mathbf{x})) \\
      CV(\boldsymbol{g}) &= mean(\boldsymbol{g})/ var(\boldsymbol{g}),
    \end{split}
    \label{eqn:diverse}
\end{equation}
where $\mathbf{x}$ is the token embedding of point $\boldsymbol{x}$, and $mean(\cdot)$ and $var(\cdot)$ compute the sample mean and variance of the input vector respectively. The diversity regularizer (\ref{eqn:diverse}) is a standard idea in MoE. Putting into a NeRF context, it encourages different views to fully exploit the expert space, and different experts to capture nuances of distinct views. 

\begin{figure*}[ht]
\centering
\vspace{-0.5em}
\begin{subfigure}[t]{0.18\textwidth}
  \centering
  \includegraphics[width=1.0\linewidth]{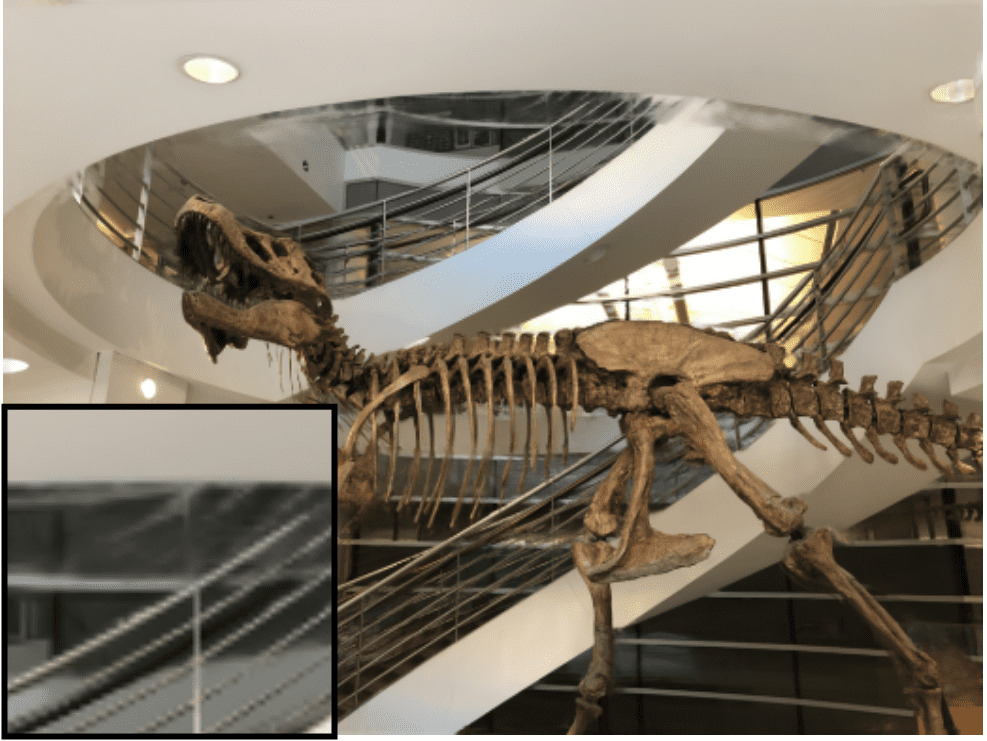}
\label{fig:trex_ibr} 
\end{subfigure}
\begin{subfigure}[t]{0.18\textwidth}
  \centering
  \includegraphics[width=1.0\linewidth]{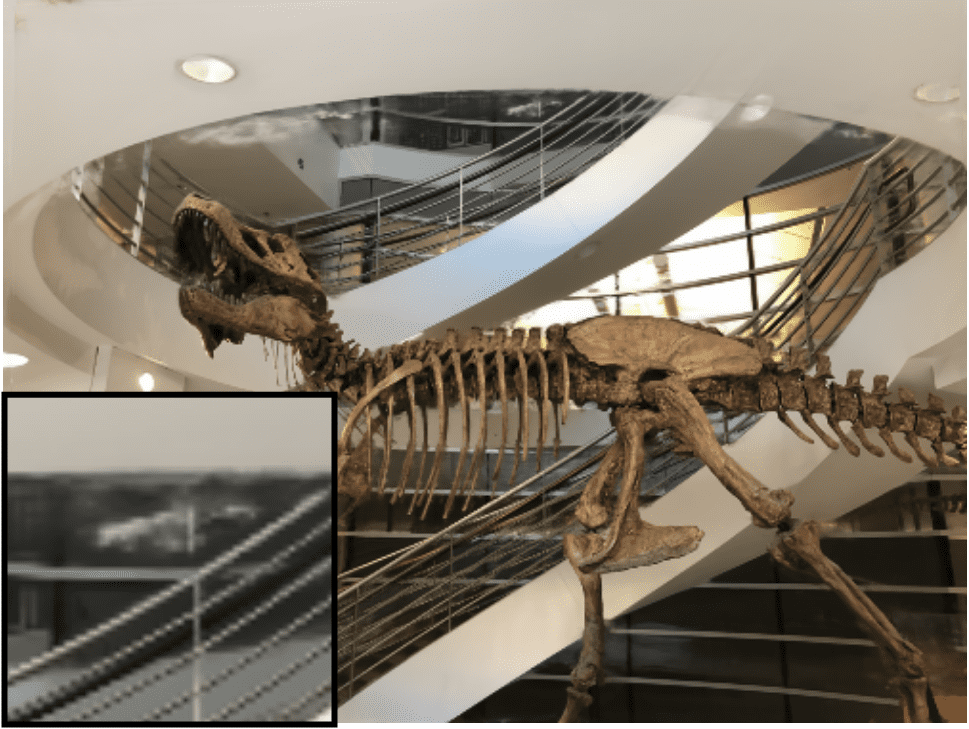}
\label{fig:trex_gnt} 
\end{subfigure}
\begin{subfigure}[t]{0.18\textwidth}
  \centering
  \includegraphics[width=1.0\linewidth]{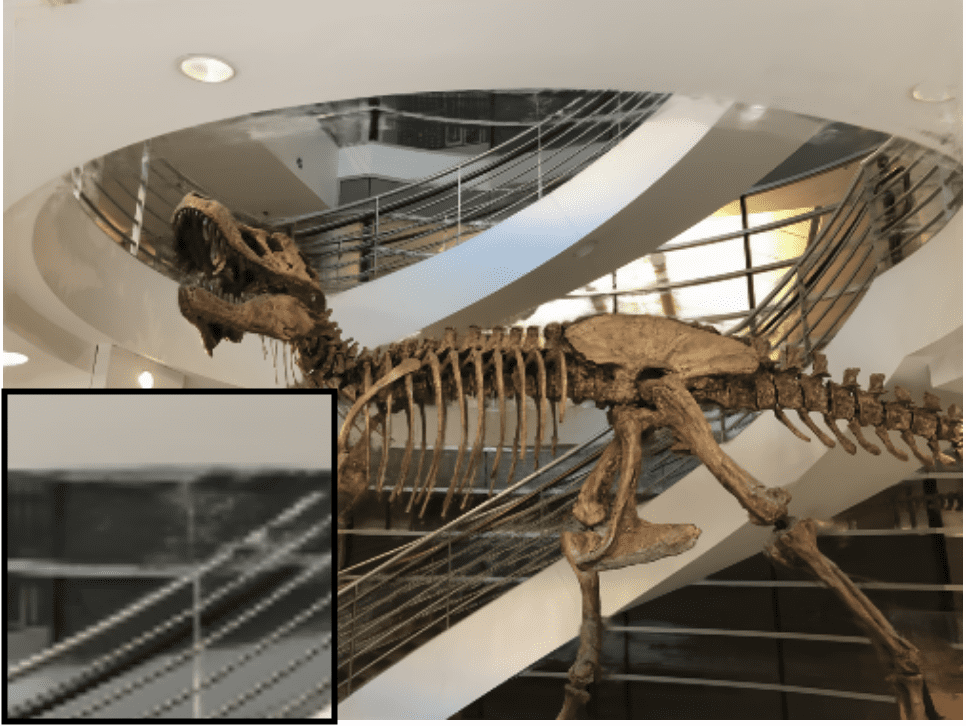}
\label{fig:trex_gpnr} 
\end{subfigure}
\begin{subfigure}[t]{0.18\textwidth}
  \centering
  \includegraphics[width=1.0\linewidth]{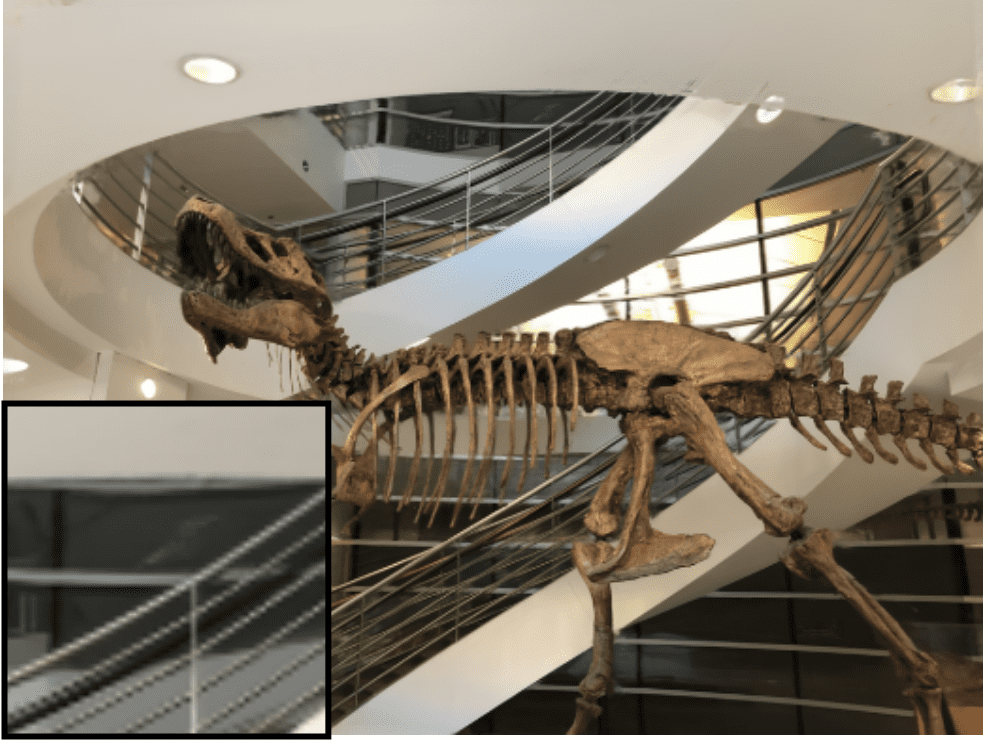}
\label{fig:trex_gntmoe} 
\end{subfigure}
\begin{subfigure}[t]{0.18\textwidth}
  \centering
  \includegraphics[width=1.0\linewidth]{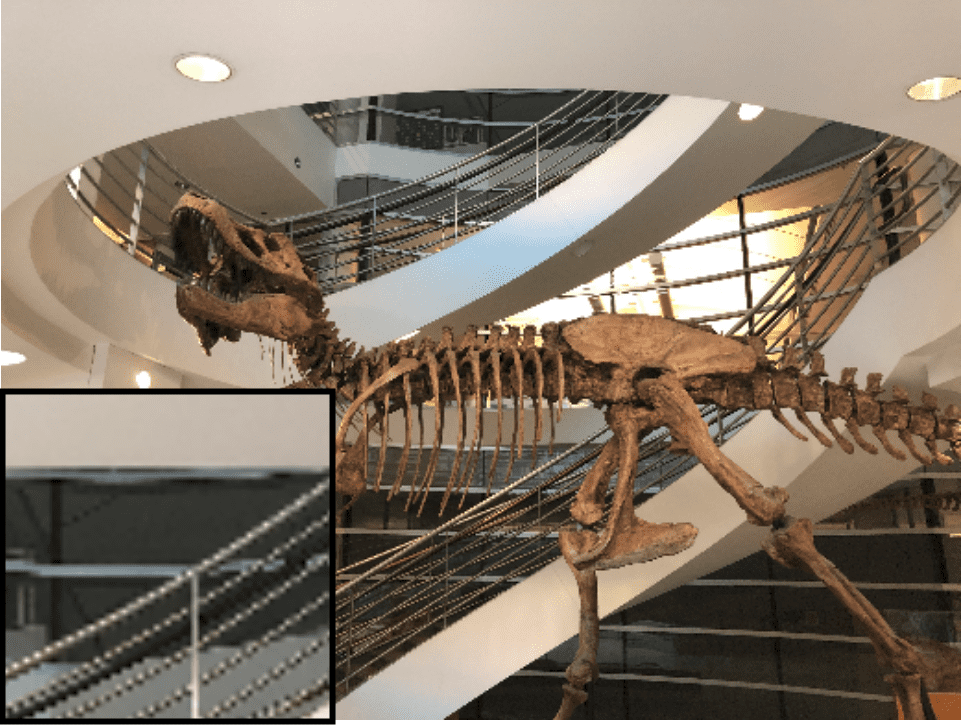}
\label{fig:trex_gt} 
\end{subfigure}

\setcounter{subfigure}{0}
\begin{subfigure}[t]{0.18\textwidth}
  \centering
  \includegraphics[width=1.0\linewidth]{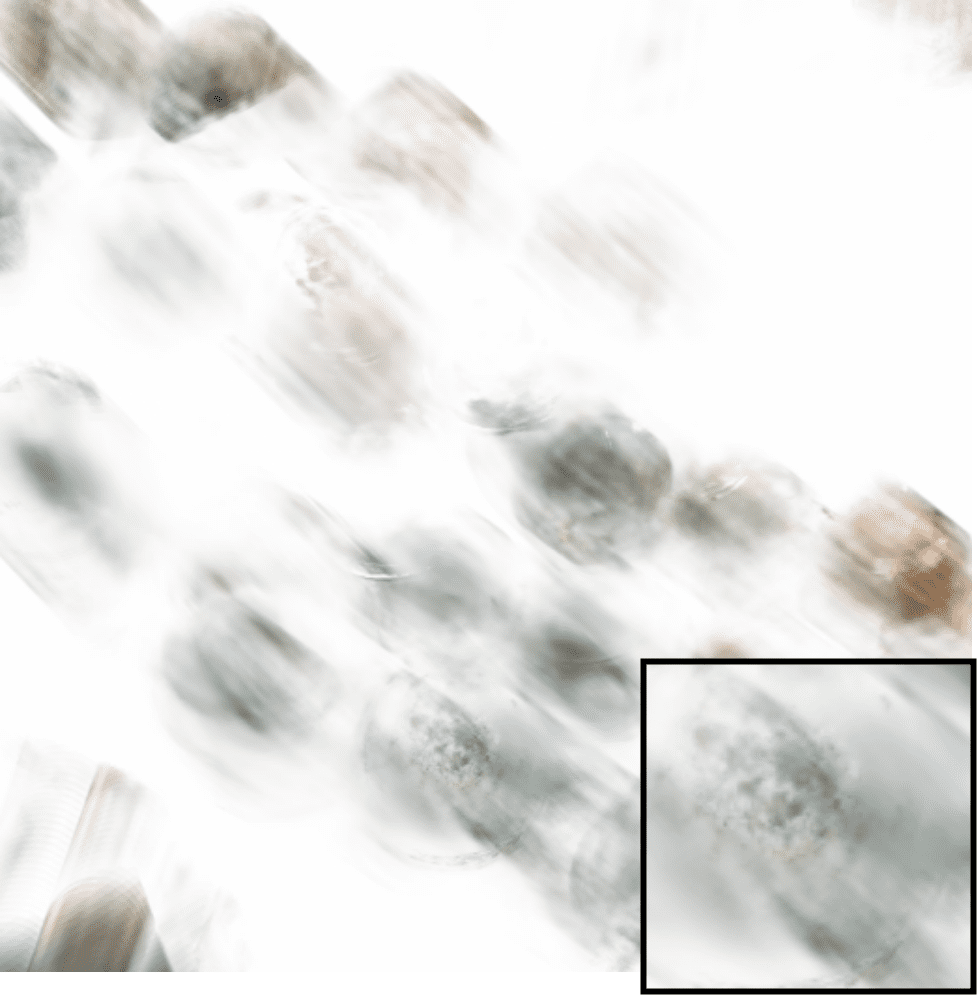}
\caption{IBRNet}
\label{fig:materials_ibr} 
\end{subfigure}
\begin{subfigure}[t]{0.18\textwidth}
  \centering
  \includegraphics[width=1.0\linewidth]{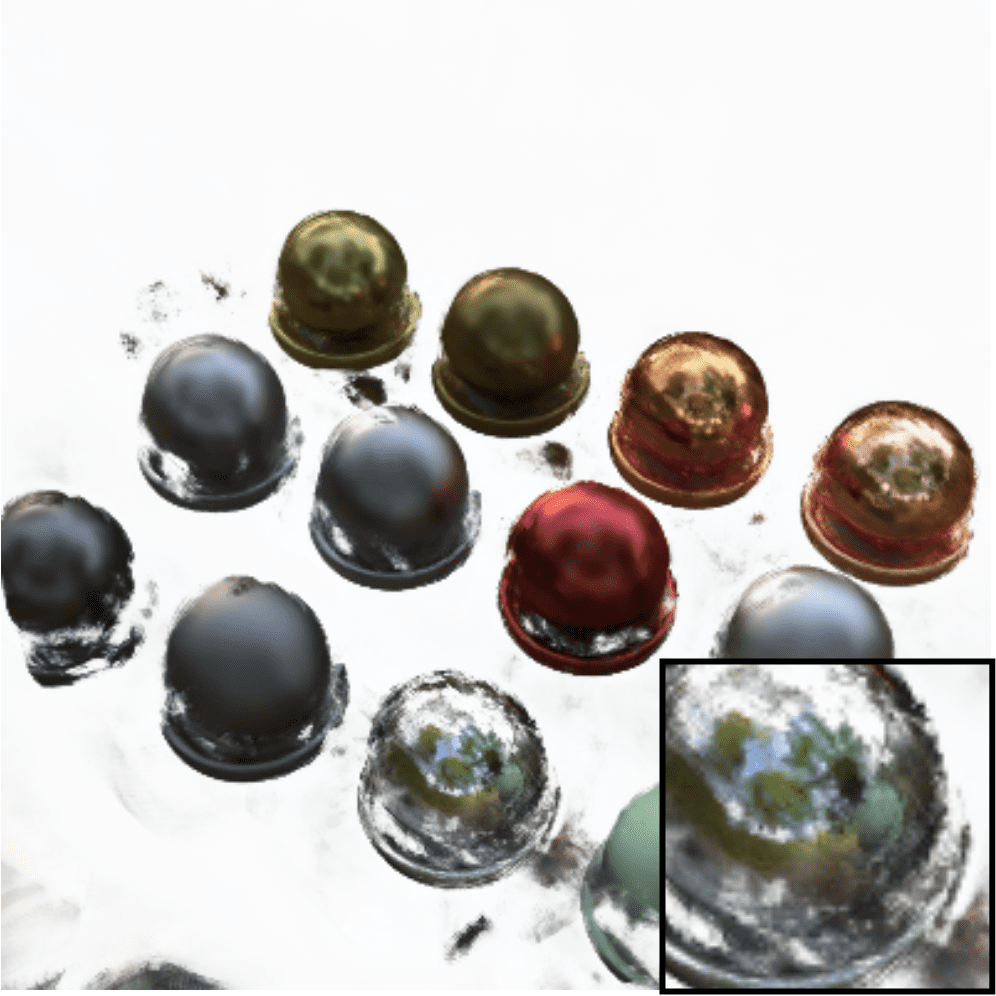}
\caption{GNT}
\label{fig:materials_gnt} 
\end{subfigure}
\begin{subfigure}[t]{0.18\textwidth}
  \centering
  \includegraphics[width=1.0\linewidth]{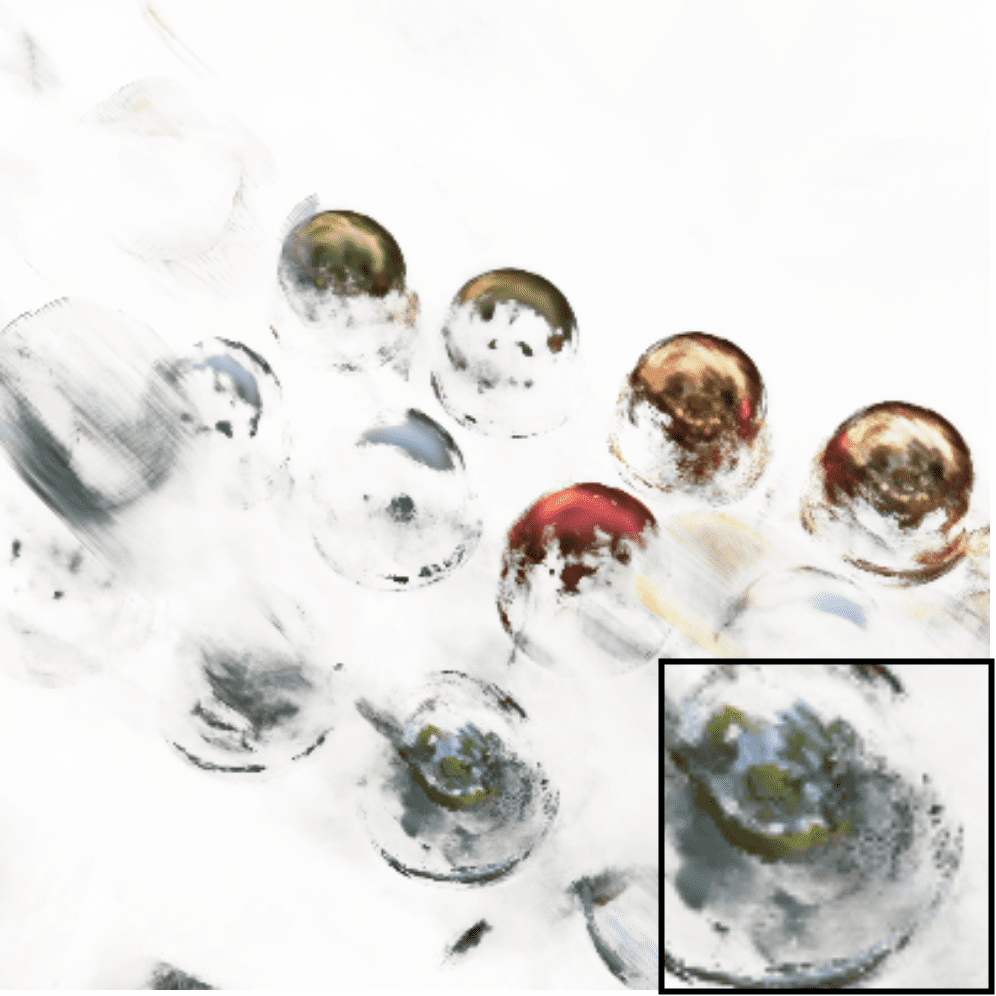}
\caption{GPNR}
\label{fig:materials_gpnr}
\end{subfigure}
\begin{subfigure}[t]{0.18\textwidth}
  \centering
  \includegraphics[width=1.0\linewidth]{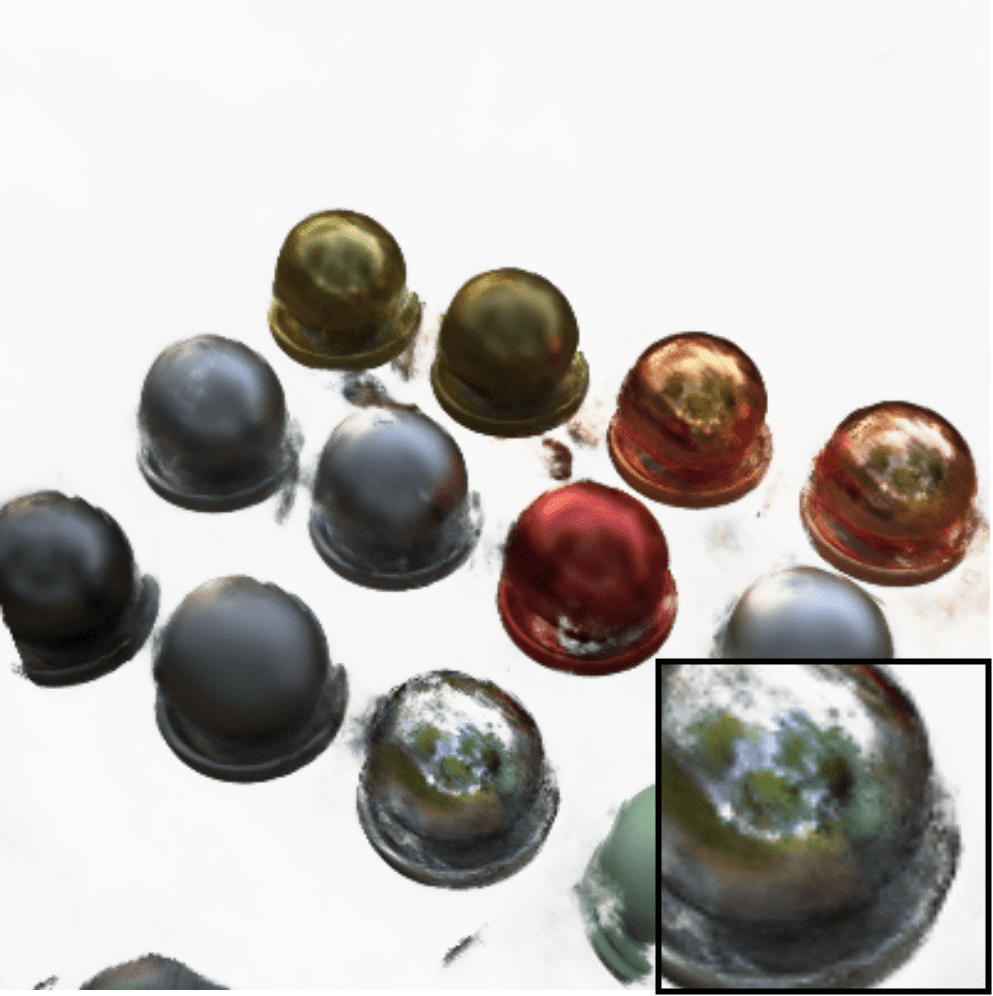}
\caption{GNT-MOVE}
\label{fig:materials_gntmoe} 
\end{subfigure}
\begin{subfigure}[t]{0.18\textwidth}
  \centering
  \includegraphics[width=1.0\linewidth]{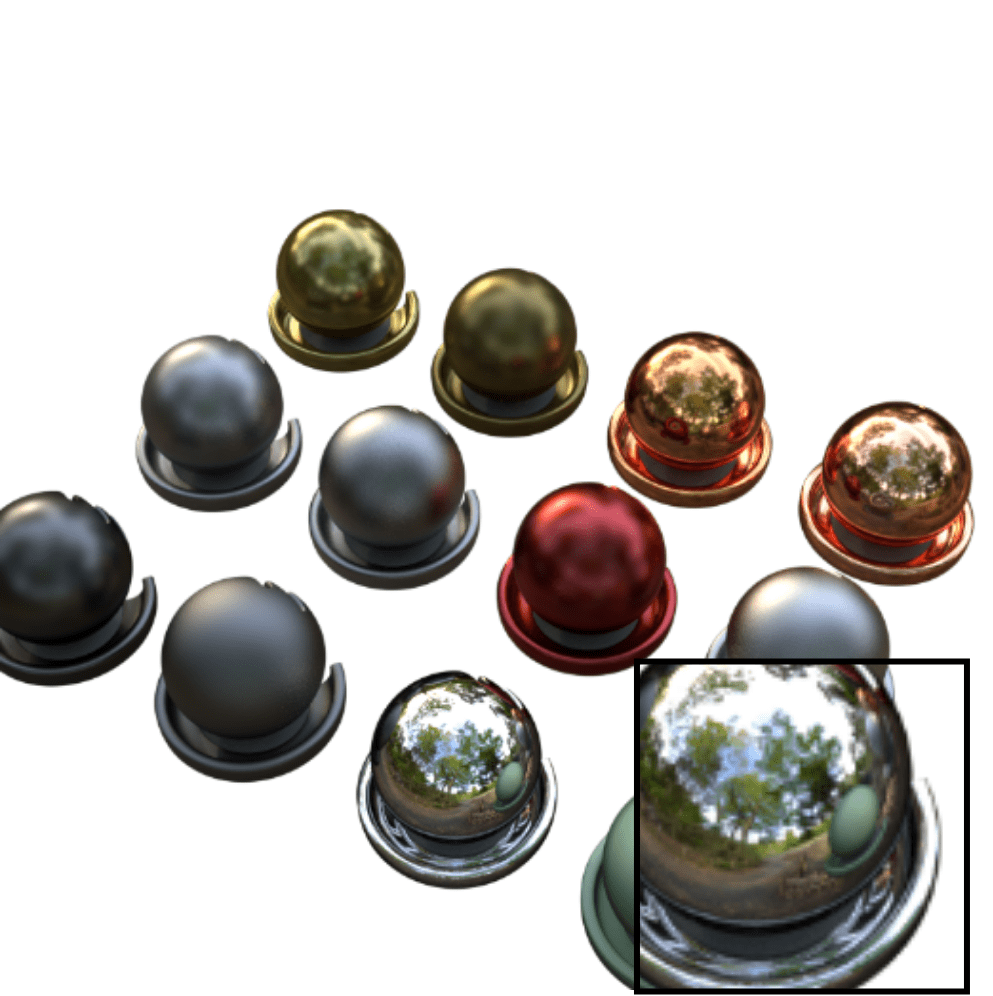}
\caption{Ground Truth}
\label{fig:materials_gt} 
\end{subfigure}
\caption{Qualitative results for the unseen cross-scene rendering. In the T-Rex scenes (row 1), GNT-MOVE reconstructs the edge details of stairs more accurately. In the Materials scenes (row 2), GNT-MOVE models the complex lighting effects much clearer compared to other methods, showing its stronger generalization ability in modeling different complex scenes. }
\label{fig:quality_result}
\vspace{-2mm}
\end{figure*}

\begin{figure}[tp]
    \begin{centering}
        \includegraphics[width=\linewidth]{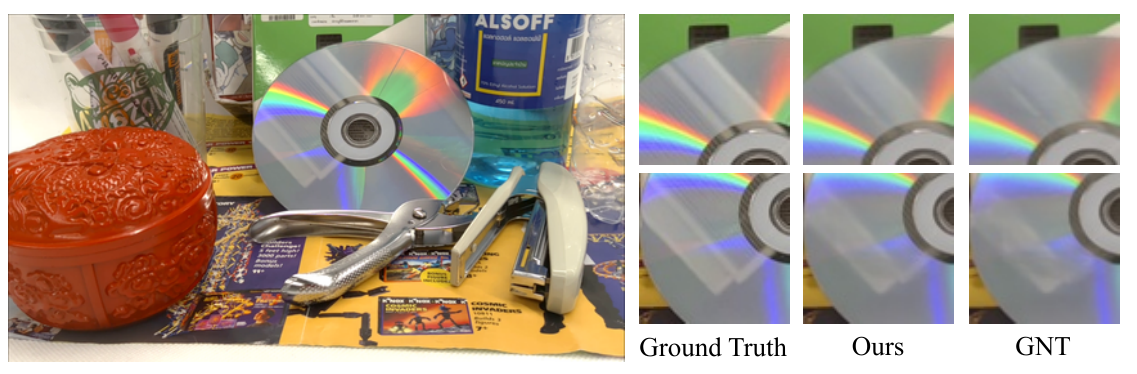}
        \par
    \end{centering}
    \caption{Qualitative comparison on Shiny-6 dataset. From left to right are the ground truth image, and the zoom-in results of GNT-MOVE and GNT, respectively}    
    \label{fig:shiny}
    \vspace{-2mm}
\end{figure}

\begin{figure}[tp]
    \begin{centering}
        \includegraphics[width=\linewidth]{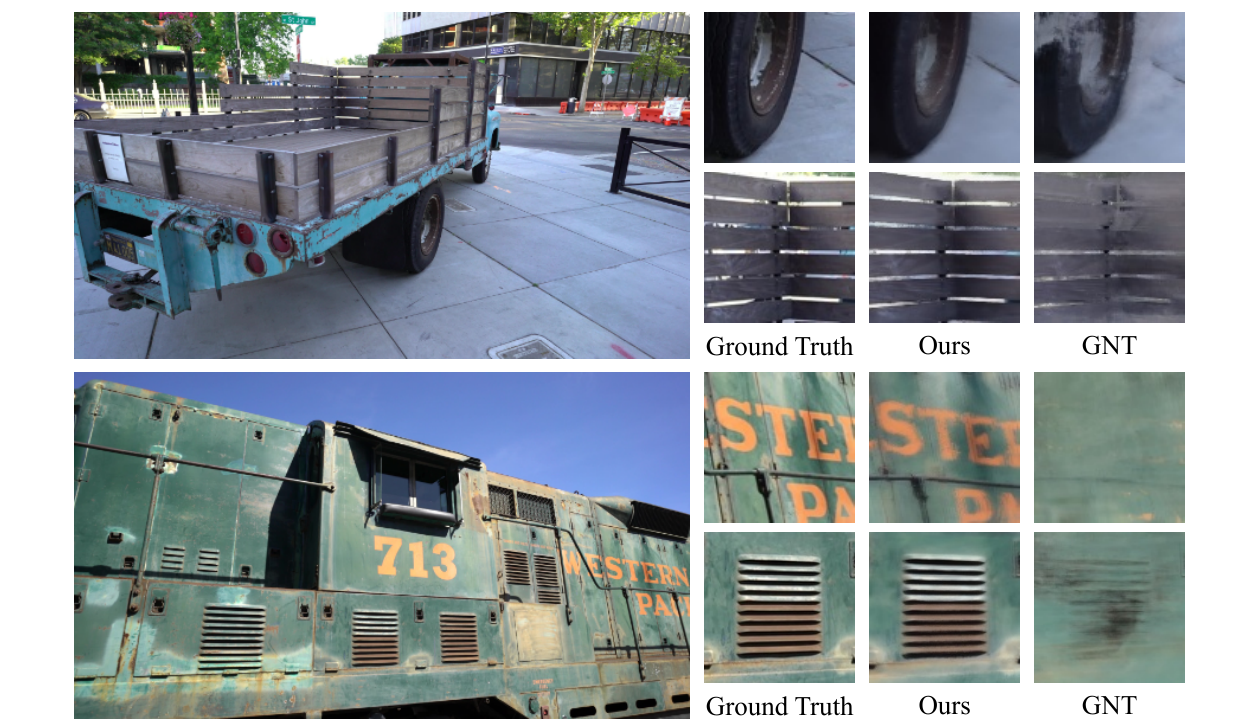}
        \par
    \end{centering}
    \caption{Qualitative comparison on Tanks-and-Temples dataset. From left to right are the ground truth image, and the zoom-in results of GNT-MOVE and GNT, respectively}   
    \label{fig:tank}
    \vspace{-2mm}
\end{figure}

\begin{figure}[tp]
    \begin{centering}
        \includegraphics[width=\linewidth]{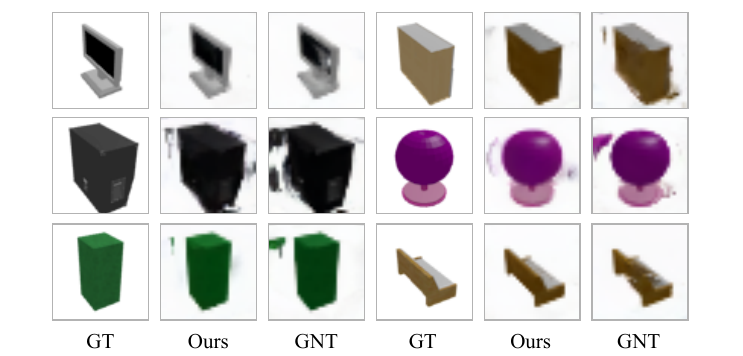}
        \par
    \end{centering}
    \caption{Qualitative comparison on NMR dataset. From left to right are the ground truth image, and the zoom-in results of GNT-MOVE and GNT, respectively.}   
    \label{fig:nmr}
    \vspace{-2mm}
\end{figure}

\subsection{Fusing Cross-scene Consistency and Spatial Smoothness into MoE} \label{sec:regularizer}

However, our experiments show that naively plugging MoE into NeRF cannot guarantee a good balance between cross-scene generalization and per-scene specialization. This is due the absence of cross-scene consistency and spatial smoothness, which are essential priors for generalizable NeRF. We hence introduce two levels of NeRF-specific customizations for MoE: (i) \underline{architecture level}: a shared permanent expert responsible for cross-scene consistency, and (ii) \underline{objective level}: a spatial consistency objective to encode geometric-aware smoothness. 

\paragraph{Permanent Shared Expert}

As aforementioned, for generalizable NeRF trained on complex and diverse scenes, the employed MoE should keep consistent expert selection on similar appearance patterns or similar materials from different scenes. However, this cross-scene consistency for NeRF can be affected by diversified expert usage in MoE. 
When we directly plug the MoE layers into GNT, we observe an obvious cross-scene inconsistency: as shown in Figure \ref{fig:permemant_expert}. For similar colors or materials from different scenes, the router selects totally different experts (\emph{e.g.}, leaves in the left sub-figure, white background in the right sub-figure), without considering the sensible cross-scene commodity. 

Therefore, to enforce said commodity across scenes and improve cross-scene consistency, we propose to modify the MoE layer from an architectural level. This is achieved through the introduction of a shared permanent expert $f_{p}$ responsible for distilling common knowledge across different scenes. The permanent expert has the same structure as other experts in the MoE. As shown in Figure \ref{fig:framework}, instead of being selected by the router, it is fixed and participates in the token processing by default.
Formally, given an input token $\mathbf{x}$ to the MoE layer, the output $\mathbf{y}$ is computed as:
\begin{equation}
 \mathbf{y} =f_{p}(\mathbf{x})+\sum_{e=1}^{E}\mathcal{R}(\mathbf{x})_{e}\cdot f_{e}(\mathbf{x})
\end{equation}

\paragraph{Geometry-Aware Spatial Consistency}
Along with the cross-scene consistency, spatial smoothness is another essential characteristic for NeRFs due to the view geometry constraints. Seeing from different camera poses, the nearby views in the same scene should make a similar or smoothly transiting expert selection. To encourage such a multi-view consistency, we propose a spatial consistency  objective that encourages two spatially close points to choose similar experts, and we use the geometric distance between them to re-weight the expert selection.

Specifically, given two spatially close 3D points $\boldsymbol{x}_i$ and $\boldsymbol{x}_j$, the router $\mathcal{R}$ takes their token embedding $\mathbf{x}_i$ and $\mathbf{x}_j$ as input and maps them to expert selection scores $R(\mathbf{x}_i), R(\mathbf{x}_j)\in \mathbbm{R}^{E}$ respectively. Similar expert selection is thereby encouraged through \textit{pulling} these two distributions closer. However, as we have a huge amount of sampled points from multiple views, it is computationally expensive and inefficient to calculate the pairwise distance between all 3D points. To make it easier to find pairs of close points, we first calculate the pairwise distance between rays based on their location in the image coordinate system. Then we filter out close rays whose pairwise distance is smaller than a predefined threshold $\epsilon$. For 3D points sampled from two close rays, we compute the Euclidean distance between all the points, denoted as $d_{i,j} = \left\|\boldsymbol{x}_i-\boldsymbol{x}_j\right\|$. For each point $\boldsymbol{x}_i$, we select its closest points $\boldsymbol{x}_i^\prime$ with distance $d_{i,i^\prime}$. 
Therefore, we encourage the consistency of the expert selection between the closest points via a symmetric Kullback–Leibler divergence loss:
\vspace{-1mm}
\begin{equation} \label{reg}
\mathcal{L}_{KL}(\boldsymbol{x}_i)=\frac{1}{2} D_{KL}(R(\mathbf{x}_i) \| R(\mathbf{x}_i^{\prime}) )+\frac{1}{2} D_{KL}(R(\mathbf{x}_i^{\prime}) \| R(\mathbf{x}_i)).
\end{equation}

As closer points are more likely to have higher expert selection similarity, we do not treat all pairs equally. Rather we use their geometric distances to serve as a consistency confidence $\rho_i=\frac{e^{-d_{i, i^\prime}}}{\sum_{(\boldsymbol{x}_j, \boldsymbol{x}_j^\prime)} e^{-d_{j, j^\prime}}}$. The final spatial consistency loss is hence defined as:

\vspace{-1mm}
\begin{equation}
\mathcal{L}_{\text {sc}}=\sum_{(\boldsymbol{x}_i, \boldsymbol{x}_i^\prime)} \rho_i \mathcal{L}_{KL}(\boldsymbol{x}_i).
\end{equation}
Note that our spatial consistency is enforced on 3D points from multiple views. Therefore, it naturally encourages geometry-aware spatial smoothness in the same scene. 

\begin{table*}[ht]
\centering
    \begin{subtable}[t]{0.43\textwidth}
    \centering
    \resizebox{\columnwidth}{!}{
        \begin{tabular}{lcccc|cccccccc}
            \toprule[1.2pt]
            \multirow{2}{*}{Models} & \multicolumn{4}{c|}{Local Light Field Fusion (LLFF)}  &\multicolumn{4}{c}{NeRF Synthetic}\\
            \cmidrule(r){2-9}
            & PSNR$\uparrow$ & SSIM$\uparrow$ & LPIPS$\downarrow$ & Avg$\downarrow$ & PSNR$\uparrow$ & SSIM$\uparrow$ & LPIPS$\downarrow$ & Avg$\downarrow$\\
            \midrule[0.8pt]
            PixelNeRF & 18.66 & 0.588 & 0.463 & 0.159 & 22.65 & 0.808 & 0.202 & 0.078\\
            MVSNeRF & 21.18 & 0.691 & 0.301 & 0.108 & 25.15 & 0.853 & 0.159 & 0.057\\
            IBRNet & 25.17 & 0.813 & 0.200 & 0.064 & 26.73 & 0.908 & 0.101 & 0.040\\
            GPNR & 25.72 & \textbf{0.880} & 0.175 & 0.055 & 26.48 & \textbf{0.944} & 0.091 & 0.036\\
            GNT & 25.86 & 0.867 & 0.116 & 0.047 & 27.29 & 0.937 & 0.056 & 0.029 \\
            \midrule
            Ours & \textbf{26.02} & 0.869 & \textbf{0.108} & \textbf{0.043} &\textbf{27.47}  & 0.940 & \textbf{0.056}& \textbf{0.029}\\
            \bottomrule[1.2pt]
        \end{tabular}
    } %
    \vspace{-1.5mm}
    \caption{NeRF Synthetic dataset and LLFF dataset.}
    \label{tab:zeroshot} 
  \end{subtable}
  \begin{subtable}[t]{0.245\textwidth}
  \centering
    \resizebox{\columnwidth}{!}{
      \begin{tabular}{llcccc}
        \toprule[1.5pt]
        \multirow{2}{*}{Setting} & \multirow{2}{*}{Models} & \multicolumn{4}{c}{Shiny-6 Dataset} \\
        \cmidrule(r){3-6}
        & & PSNR$\uparrow$ & SSIM$\uparrow$ & LPIPS$\downarrow$ & Avg$\downarrow$ \\
        \midrule[1pt]
        \multirow{4}{*}{\makecell{Per-Scene \\ Training}} & NeRF & 25.60 & 0.851 &  0.259 & 0.065 \\
        & NeX & 26.45 & 0.890 & 0.165 & 0.049 \\
        & IBRNet & 26.50 &  0.863 & 0.122 & 0.047\\
        & NLF & 27.34 & 0.907 &\textbf{ 0.045 }& \textbf{0.029}\\
        \midrule
        \multirow{4}{*}{Generalizable} & IBRNet & 23.60 & 0.785 & 0.180 & 0.071 \\
        & GPNR & 24.12 & 0.860 & 0.170 & 0.063\\
        \cmidrule(r){2-6}
        & GNT & 27.10 & 0.912 & 0.083 & 0.036 \\
        & Ours & \textbf{27.54} & \textbf{0.932} & 0.072 & 0.032 \\
        \bottomrule[1.5pt]
      \end{tabular}}
    \vspace{-1.5mm}
    \caption{Shiny dataset.}
    \label{tab:zeroshot_shiny}
  \end{subtable}
  \begin{subtable}[t]{0.31\textwidth}
    \centering
    \resizebox{\columnwidth}{!}{
    \begin{tabular}{lcccc}
     \toprule
     \multirow{2}{*}{Models} & \multicolumn{4}{c}{NMR Dataset}  \\
        \cmidrule(r){2-5}
      & PSNR $\uparrow$ & SSIM $\uparrow$ & LPIPS $\downarrow$ & Avg $\downarrow$ \\
    \midrule
    LFN & 24.95 & 0.870 & - & - \\
    PixelNeRF & 26.80 & 0.910 & 0108 & 0.041 \\
    SRT & 27.87 & 0.912 & 0.066 & 0.032 \\
    \hline GNT & 32.12 & 0.970 & 0.032 & 0.015 \\
    Ours & \textbf{33.08} & \textbf{0.972} & \textbf{0.031} & \textbf{0.014} \\
    \bottomrule
    \end{tabular}
    }
    \vspace{-1.5mm}
    \caption{NMR dataset.}
    \label{tab:nmr}
\end{subtable}

\begin{subtable}[t]{0.9\textwidth}
\centering
\resizebox{\columnwidth}{!}{
\begin{tabular}{llcccccccccccc}
\toprule
 \multirow{2}{*}{Setting} & \multirow{2}{*}{Models}  & \multicolumn{3}{c}{Truck} & \multicolumn{3}{c}{Train} & \multicolumn{3}{c}{M60} & \multicolumn{3}{c}{Playground} \\
 \cmidrule(r){3-14}
    &  &   PSNR$\uparrow$ & SSIM$\uparrow$ & LPIPS$\downarrow$  &    PSNR$\uparrow$ & SSIM$\uparrow$ & LPIPS$\downarrow$ &     PSNR$\uparrow$ & SSIM$\uparrow$ & LPIPS$\downarrow$    &     PSNR$\uparrow$ & SSIM$\uparrow$ & LPIPS$\downarrow$    \\ \midrule
\multirow{2}{*}{Per-scene Training} &  NeRF & 20.85 & 0.747 & 0.513 & 16.64 & 0.635  & 0.651 & 16.86 & 0.702 & 0.602 & 21.55  & 0.765 & 0.529 \\
    & NeRF++ &  22.77  & 0.823 & 0.298  & 17.17 & 0.672  & 0.523  & 17.88 & 0.738   & 0.435     & 22.37 &  0.799 & 0.391   \\ \hline
     \midrule
\multirow{2}{*}{Generalizable} & GNT  &  17.39  &    0.561   &  0.429    &   14.09    &    0.420   &   0.552   &     11.29     &  0.419   &   0.605   &   15.36   &   0.417    &   0.558 \\  
   & Ours   & 19.71  &  0.628   &  0.379 &   16.27     &  0.499     &   0.466 &    13.56  &    0.495    &   0.527    &   19.10   & 0.501  & 0.507   \\  
\bottomrule
\end{tabular}
}
\vspace{-1.5mm}
\caption{Tanks-and-Temples dataset.}
\label{tab:tnt}
\end{subtable}
\vspace{-1.5mm}
\caption{Comparison of GNT-MOVE against SOTA methods for cross-scene generalization under \textbf{zero-shot setting}.}
\vspace{-1.5mm}
\end{table*}

\section{Experiments}
In this section, we conduct extensive experiments with GNT-MOVE to answer two questions: 
\textit{i) Does MoE help GNT scale up in scene coverage and improve generality?}
\textit{ii) Does GNT-MOVE meanwhile improve specialization to different scenes?}
We compare GNT-MOVE with state-of-the-art (SOTA) methods on generalizable novel view synthesis tasks, under both zero-shot and few-shot settings (Section \ref{sec:expr_generalization}).
We also provide careful analyses on the expert selection in GNT-MOVE to illustrate how MoE divide and conquer to render a challenging scene (Section \ref{sec:expr_expert_analysis}).

\subsection{Implementation Details}
\paragraph{Training / Inference Details}
We choose top $K=2$ experts out of $E=4$ expert candidates per layer. Note that we scale down the expert size by half compared to the dense MLP layer in standard ViT to make their computation FLOPs equivalent. We train GNT-MOVE end-to-end using the Adam optimizer. The threshold $\epsilon$ for close rays is set as 20. The loss weights $\lambda_{sc}$ and $\lambda_{div}$ are set to be $1\times 10^{4}$ and $1\times 10^{-3}$, respectively. Please refer to our supplementary for additional training details. 

\vspace{-3mm}
\paragraph{Metrics} 
We adopt three widely-used metrics: Peak Signal-to-Noise Ratio (PSNR), Structural
Similarity Index Measure (SSIM)~\cite{wang2004image}, and the Learned Perceptual Image Patch Similarity (LPIPS)~\cite{zhang2018unreasonable}. We report the average of each metric across multiple scenes in one dataset for cross-scene generalization experiments. Following \cite{gnt}, we also report the geometric mean of $10^{-PSNR/10}$,$\sqrt{1-SSIM}$, LPIPS, for an easier comparison~\cite{barron2021mip}.

\subsection{Main Experiments: Zero-Shot and Few-Shot Cross-Scene Generalization} \label{sec:expr_generalization}

\paragraph{Setting} To evaluate the cross-scene generalization performance, we compare our GNT-MOVE with state-of-the-art generalizable NeRF under two important settings:
\begin{itemize}\vspace{-0.3em}
    \item \textbf{Zero-shot}: the pre-trained model is directly evaluated on an unseen scene for novel view synthesis.\vspace{-0.2em}
    \item  \textbf{Few-shot}: the pre-trained model is first finetuned with a few observed views from the target unseen scene, and then applied to the target scene.\vspace{-0.3em}
\end{itemize}

\vspace{-5mm}
\paragraph{Datasets} We follow the experimental protocol in IBRNet~\cite{wang2021ibrnet} and GNT~\cite{gnt} and use the following training/evaluation datasets: (1) \textbf{Training Datasets} consist of both real and synthetic data, in consistency with GNT \cite{gnt}. For synthetic data, we use object renderings of 1023 models from Google Scanned Object~\cite{downs2022google}. For real data, we make use of RealEstate10K~\cite{zhou2018stereo}, 90 scenes from the Spaces dataset~\cite{flynn2019deepview}, and 102 real scenes from handheld cellphone captures~\cite{mildenhall2019local, wang2021ibrnet}. (2) \textbf{Testing Datasets} are the common NeRF benchmarks including Local Light Field Fusion (LLFF)~\cite{mildenhall2019local} and NeRF Synthetic dataset~\cite{NERF}. Note that these LLFF scenes are not included in the handheld cellphone captures in the training set.
We also include \textit{three additional datasets}: Shiny-6 dataset~\cite{wizadwongsa2021nex}, Tanks-and-Temples~\cite{riegler2020free} and NMR~\cite{kato2018neural}, which contains complex optical effects, large unbounded scenes, and 360\textdegree~views of various objects from unseen categories, respectively.
More dataset details can be found in the \underline{supplementary}.

\vspace{-2mm}

\subsubsection{Zero-Shot Generalization}

\begin{table*}[t]
    \centering
    \resizebox{\textwidth}{!}{
    \begin{tabular}{l|cccc|cccc|cccc|cccc|cccc}
        \toprule[1.5pt]
        \multirow{3}{*}{Models} & \multicolumn{12}{c|}{Local Light Field Fusion (LLFF)} & \multicolumn{8}{c}{NeRF Synthetic} \\
        \cmidrule{2-21}
        & \multicolumn{4}{c|}{3-shot}& \multicolumn{4}{c|}{6-shot}& \multicolumn{4}{c|}{10-shot} & \multicolumn{4}{c|}{6-shot}& \multicolumn{4}{c}{12-shot} \\
        \cmidrule{2-21}
        & PSNR$\uparrow$ & SSIM$\uparrow$ & LPIPS$\downarrow$ & Avg$\downarrow$ & PSNR$\uparrow$ & SSIM$\uparrow$ & LPIPS$\downarrow$ & Avg$\downarrow$ & PSNR$\uparrow$ & SSIM$\uparrow$ & LPIPS$\downarrow$ & Avg$\downarrow$ & PSNR$\uparrow$ & SSIM$\uparrow$ & LPIPS$\downarrow$ & Avg$\downarrow$ & PSNR$\uparrow$ & SSIM$\uparrow$ & LPIPS$\downarrow$ & Avg$\downarrow$\\
        \midrule[1pt]
        PixelNeRF & 17.54 & 0.543 & 0.502 & 0.181 & 19.00 & 0.721 & 0.496 & 0.148 & 20.01 & 0.755 & 0.333 & 0.123 & 19.13 & 0.783 & 0.250 & 0.112 & 21.90 & 0.849 & 0.173 & 0.075 \\
        MVSNeRF &17.05 & 0.486 & 0.480 &0.189 &20.50 & 0.594 & 0.384 &0.130 & 22.54 & 0.673 & 0.309 &0.099 & 16.74 & 0.781 & 0.263 &0.138 & 22.06 & 0.844 & 0.185 & 0.076\\
        IBRNet  &16.89 &0.539 &0.458 & 0.185 &20.61 &0.686 &0.316 &0.115 &23.52 &0.789 &0.226 & 0.077 & 18.17 &0.812 & 0.234 &0.115 &24.69 &0.895 &0.120 &0.051\\
        GNT     &19.58 &0.653 &0.279 &0.121 &22.36 & 0.766 &0.189 &0.081 &24.14 &0.834 & 0.133 &0.059 &22.39 &0.856 &0.139 &0.067 & 25.25 & 0.901 & 0.088& 0.044\\
        \cmidrule{1-21}
        Ours    & \textbf{19.71} &\textbf{0.666} &\textbf{0.270} &\textbf{0.120} &\textbf{22.53} &\textbf{0.774} &\textbf{0.184} & \textbf{0.078} &\textbf{24.61} &\textbf{0.837} &\textbf{0.132} & \textbf{0.056} &\textbf{22.53} &\textbf{0.871} &\textbf{0.116} & \textbf{0.061} &\textbf{25.85} &\textbf{0.915} &\textbf{0.074} & \textbf{0.038}\\
        \bottomrule[1.5pt]
    \end{tabular}
    }
    \caption{Comparison of GNT-MOVE against SOTA methods in \textbf{few-shot setting} on the LLFF and NeRF Synthetic datasets.}
    \vspace{-2mm}
    \label{tab:fewshot-llff-nerf-synthetic}
\end{table*}

For LLFF and NeRF Synthetic scenes, we compare our method with PixelNeRF~\cite{yu2021pixelnerf}, MVSNeRF~\cite{chen2021mvsnerf}, IBRNet~\cite{wang2021ibrnet}, GNT~\cite{gnt}, and GPNR~\cite{gpnr}. As seen from Table \ref{tab:zeroshot}, our method achieves the best performance on both LLFF and NeRF Synthetic datasets in PSNR, LPIPS, and average evaluation metrics. 
Compared with GPNR, GNT-MOVE achieves a significantly better perceptual score, with up to $38\%$ LPIPS reduction on both datasets. We also outperform GNT on PSNR with notable improvements of $0.16$dB and $0.18$dB on two datasets. 
The qualitative results on representative scenes are shown in Figure \ref{fig:quality_result}. One could observe that GNT-MOVE renders novel views with clearly better visual quality. It particularly better reconstructs fine details of object edges in T-Rex, and more accurately models complex specular reflection effects in Materials (even our training sets contain only limited lighting variations).

We then compare on the more challenging Shiny \cite{wizadwongsa2021nex}, Tanks-and-Temples \cite{riegler2020free}, and NMR \cite{kato2018neural} datasets. On the non-object-centric Shiny dataset, we observe from Table \ref{tab:zeroshot_shiny} that, GNT-MOVE clearly surpasses its peers of the generalizable category in all the metrics: outperforming GNT by 0.44 dB PSNR, and GPNR/IBRNet by over 3 dB. Even compared to the per-scene fitting category (which puts an unfair disadvantage on us), our PSNR and SSIM still win over strong competitors such as NLF (which has particular optical modeling) and IBRNet. All those results endorse that GNT-MOVE benefits its MoE-based specialization to adapt well to challenging materials and light effects.

We further compare GNT-MOVE with SRT \cite{sajjadi2022scene}, another transformer-based renderer pre-trained by novel view synthesis, on the NMR dataset \cite{kato2018neural}.
As shown in Table \ref{tab:nmr}, GNT-MOVE remarkably outperforms SRT by 5.21 dB PSNR - that is even more impressive if one considers that SRT is pre-trained with samples from NMR. In contrast, GNT-MOVE can ``zero-shot" generalize way better. It also outperforms GNT by a remarkable 1 dB PSNR. Table \ref{tab:tnt} also demonstrates the performance of GNT-MOVE on the Tanks-and-Temples dataset (the four scenes selected in NeRF++\cite{zhang2020nerf}). Once again, GNT-MOVE largely outperforms GNT by up to 3 dB PSNR. Those results strongly suggest that GNT-MOVE, with its higher capacity, is indeed more generalizable and robust than vanilla GNT. The qualitative rendering comparison on representative scenes from Shiny-6 dataset~\cite{wizadwongsa2021nex}, Tanks-and-Temples dataset~\cite{riegler2020free}, and NMR dataset~\cite{kato2018neural} could be found in Figure \ref{fig:shiny}, Figure \ref{fig:tank}, and Figure \ref{fig:nmr}, respectively.

\vspace{-1em}

\subsubsection{Few-Shot Generalization}

Next under the few-shot setting, we compare our method with PixelNeRF~\cite{yu2021pixelnerf}, MVSNeRF~\cite{chen2021mvsnerf}, IBRNet~\cite{wang2021ibrnet}, and GNT~\cite{gnt}. On the LLFF dataset that contains forward-facing scenes, we finetune the pre-trained models using 3, 6, and 10 images. On NeRF Synthetic dataset that contains $360^{\circ}$ scenes, we finetune them on 6 and 12 images, respectively. During inference, images used for finetuning are by default included as source images for novel view synthesis.

In Table \ref{tab:fewshot-llff-nerf-synthetic}, GNT-MOVE shows a remarkably large performance gain over all the state-of-the-art methods on NeRF Synthetic dataset. Compared to GNT, our model achieves better results in all metrics, with particularly impressive perceptual score gains of $17\%$ and $16\%$ LPIPS on 6-shot and 12-shot, respectively.  GNT-MOVE also improves over GNT by a great margin of $0.6$ dB PSNR and $0.14$ SSIM on 12-shot setting. Similar performance gains are also observed on the LLFF dataset: GNT-MOVE improves the state-of-the-art GNT on PSNR metric by $0.13$ dB, $0.17$ dB, and $0.47$ dB on 3-shot, 6-shot, and 10-shot, respectively.

\begin{figure*}[tp]
\vspace{-1mm}
    \begin{centering}
        \includegraphics[width=1.0\linewidth]{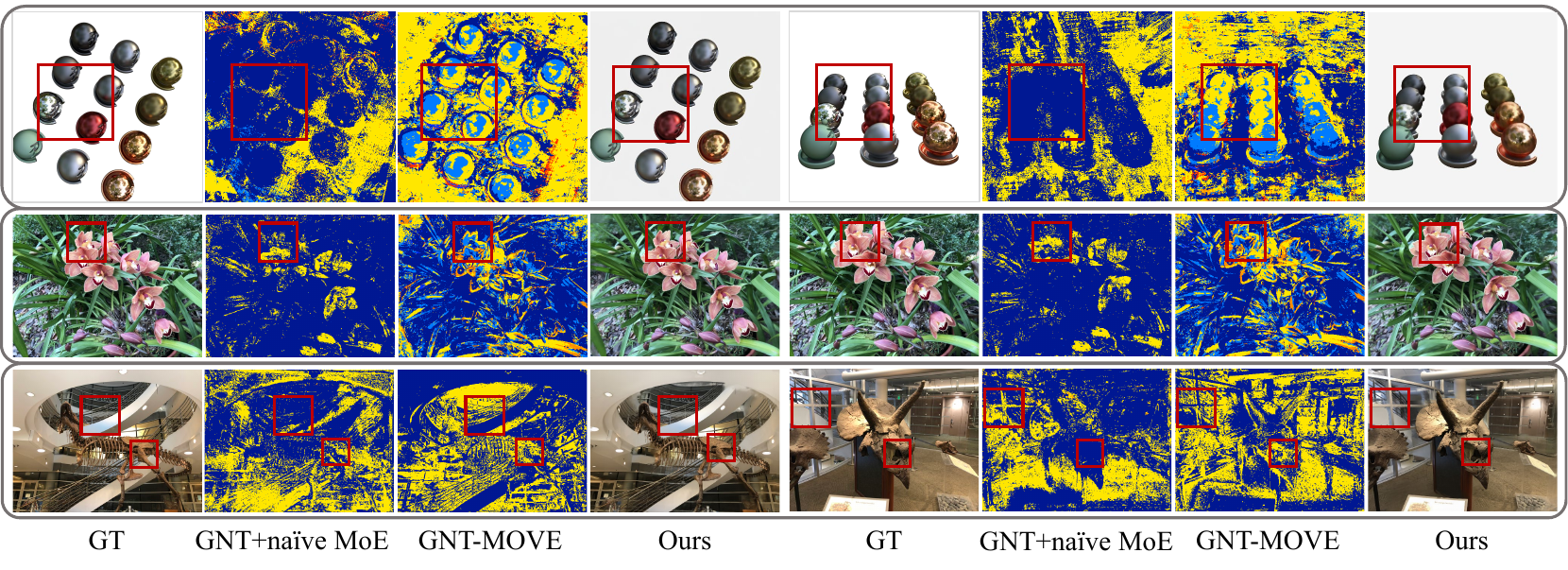}
        \par
    \end{centering}
    \vspace{-1mm}
    \caption{Visualization of expert selection using different colors. Row 1: two quite different views from the Materials scene. Row 2: two slightly different views from the Orchids scene. Row 3: two different scenes, but with similar visual appearances (\emph{e.g.}, stairs, bones). We compare GNT with naive MoE, and our GNT-MOVE solution.}    
    \label{fig:vis_expert}
    \vspace{-2mm}
\end{figure*}

\subsection{Spotlight Comparison: GNT v.s. GNT-MOVE}

Since GNT-MOVE is an extension of GNT (which is the most recent SOTA), it is naturally of interest to compare the two closely and to understand how much benefits MoE actually brings to GNT (``specialization" v.s. ``generalization" ), for the goal of cross-scene generalization. While most aforementioned experiments already demonstrate various solid gains, we feel it worthy of providing a focused summary below. We emphasize that GNT and GNT-MOVE are trained and evaluated in completely fair settings.\vspace{-1mm}
\begin{itemize}
    \item In the zero-shot setting, GNT-MOVE always outperforms GNT on the metric of PSNR, with moderate improvements of \textbf{0.16 dB} and \textbf{0.18 dB}, on the ``standard" LLFF and NeRF Synthetic datasets, respectively. Yet on the more challenging ones, the PSNR gain of GNT-MOVE over GNT becomes larger: \textbf{0.44 dB} on Shiny, \textbf{0.96 dB} on NMR, and eventually an impressive \textbf{2.63 dB} on Tanks-and-Temples (averaged over 4 scenes).\vspace{-1mm} 
    \item Same in the zero-shot setting, GNT-MOVE outperforms GNT in all cases on the metrics of LPIPS and Avg scores. It marginally lags behind GPNR on SSIM in NeRF Synthetic and LLFF, but wins on SSIM on other more challenging datasets. For example, the SSIM gain of GNT-MOVE over GNT is as large as 0.076 on Tanks-and-Temples (averaged over 4 scenes).\vspace{-1mm} 
    \item Then, in the few-shot setting, our results suggest a \textbf{clean sweep} for GNT-MOVE, in all shot settings, under all metrics, on both LLFF and NeRF Synthetic datasets. Generally, as the number of shots increases, the gains of GNT-MOVE over GNT seem to increase as well, ending up with \textbf{0.47 dB} and \textbf{0.60 dB} gaps on LLFF and NeRF synthetic, respectively.\vspace{-1mm}
    \item When it comes to visual quality, GNT-MOVE is clearly superior in tackling challenging scenes with complex lighting, e.g., Ship, Materials, and Drums (please refer to the per-scene breakdown results of zero-shot generalization in the \underline{supplementary}). The experiments on the Shiny dataset in Table~\ref{tab:zeroshot_shiny} demonstrate that GNT-MOVE generalizes better than GNT in the presence of challenging refraction and reflection.\vspace{-1mm} 
    \item Also in Table \ref{tab:tnt}, GNT-MOVE generalizes out of the box on large-scale, unbounded 3D scenes while the vanilla GNT fails. Note that both GNT-MOVE and GNT are trained only on bounded and forward-facing scenes, implying the stronger compositional generalization potential \cite{keysersmeasuring} achieved through MoEs.\vspace{-1mm}
\end{itemize}

More comparisons, demonstrating that the solid gain of MoE for generalizable NeRF goes way beyond naively larger model size; yet the gain can only be unleashed with PE and SR, could be found in the \underline{supplementary}.

\begin{figure}[t]
\centering
\vspace{-1mm}
\subfloat[]{
\label{fig:cross-scene}
\includegraphics[width=0.28\linewidth]{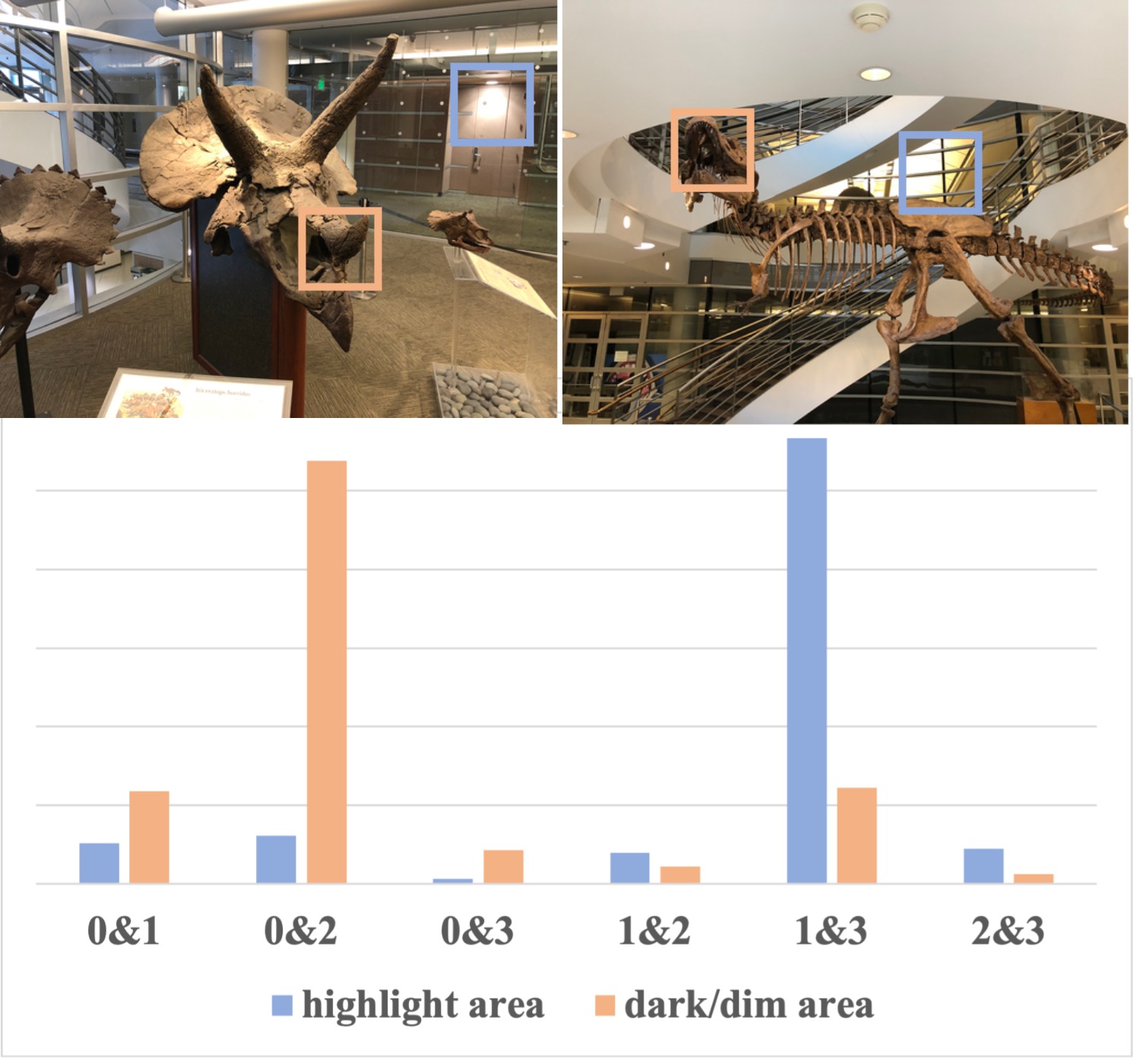}
\vspace{-1mm}}
\subfloat[]{
\label{fig:cross-view}
\includegraphics[width=0.62\linewidth]{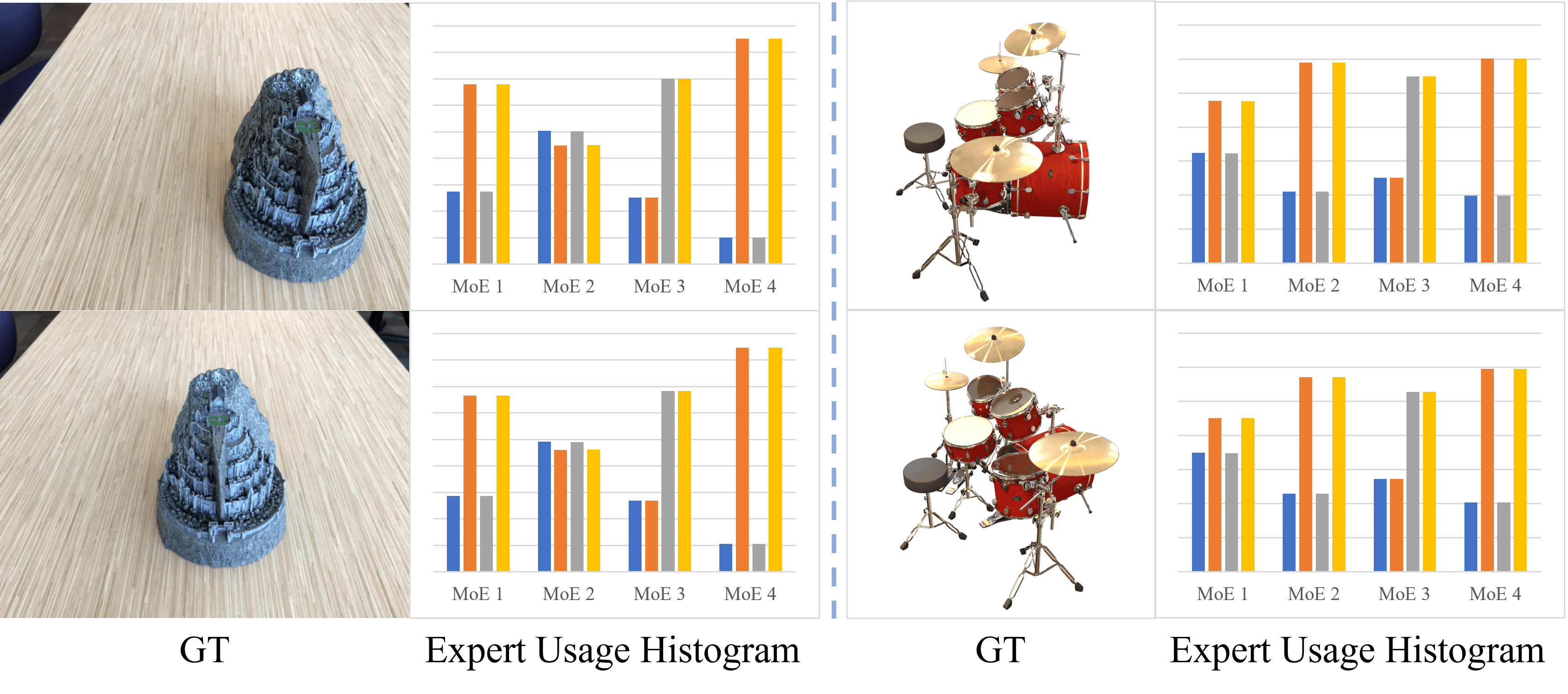}\vspace{-2mm}}
\quad\vspace{-2mm}
\caption[]{Expert selection histogram. (a) Similar patterns (e.g., bright) across different scenes have similar expert selections; (b) Different views from the same scene have similar and consistent layer-wise expert selections.}
\label{fig:exp-sel-his}
\vspace{-2mm}
\end{figure}

\begin{table*}[htb]
\centering
\resizebox{0.75\linewidth}{!}{
\begin{tabular}{lccc|cccc|cccccccc}
\toprule
\multicolumn{4}{c|}{Models} &  \multicolumn{4}{c|}{Local Light Field Fusion (LLFF)}  &\multicolumn{4}{c}{NeRF Synthetic}\\
\cmidrule(r){1-12}
&MoE& PE & SR  & PSNR$\uparrow$ & SSIM$\uparrow$ & LPIPS$\downarrow$ & Avg$\downarrow$ & PSNR$\uparrow$ & SSIM$\uparrow$ & LPIPS$\downarrow$ & Avg$\downarrow$\\
\midrule
GNT &--&--&--& 25.86 & 0.867 & 0.116 & 0.047 & 27.29 & 0.937 & 0.056 & 0.029 \\
\midrule
Ours &\ding{51}&&& 25.46 & 0.856 & 0.128 &0.051   & 27.15 & 0.934 & 0.057 &  0.031 \\

Ours &\ding{51}&\ding{51}&& 25.88 & 0.865 &0.120 &  0.049  &27.32  & 0.936 & 0.058 &0.030  \\
Ours &\ding{51}&&\ding{51}& 25.93 & 0.866 & 0.117 & 0.046  &27.30  & 0.935 & 0.059 & 0.030 \\
Ours &\ding{51}&\ding{51}&\ding{51}& \textbf{26.02} & \textbf{0.869} & \textbf{0.108} & \textbf{0.043} &\textbf{27.47}  & \textbf{0.940} & \textbf{0.056}& \textbf{0.029}\\
\bottomrule
\end{tabular}
}
\caption{Ablation analyses of our two key proposals: PE indicates permanent expert and SR indicates smoothness regularizer.} %
\label{tab:ablation} 
\end{table*}

\subsection{Dive into the Expert Selection} \label{sec:expr_expert_analysis}

GNT-MOVE strikes a good balance between cross-scene/view consistency and expert specialization: it can be demonstrated through the visualization of expert maps in Figure \ref{fig:vis_expert}, where we compare GNT-MOVE with a baseline of GNT + naive MoE (i.e., the basic pipeline described in Sec. \ref{sec:pipeline}, without enforcing our customized consistency/smoothness). 

In Row 1, two different views of the Materials scene select the same set of experts for foreground material balls and the background, respectively. That is in contrast to the much more confused/``mixed" selection observed in GNT + naive MoE. In Row 2, one observes the same cross-view consistency, while the subtle differences between two views (\emph{e.g.}, occluded bud) are also modeled differently in the two corresponding expert maps, indicating good expert specialization and diversity. Row 3 indicates an example of cross-scene consistency, where the same expert group is selected by GNT-MOVE for similar visual appearances (\emph{e.g.}, stairs, bones) across two different scenes.

The selection of experts also properly reacts to fine edges (\emph{e.g.}, flower edges in row 2, handrail edge and bone edge in row 3), and is also capable of adapting to complex lighting effects, as shown in the Materials scene  (row 1) and the light part of the T-Rex scene (row 3 right). Furthermore, we visualize the expert selection histogram in Figure \ref{fig:exp-sel-his}. It aligns well with our observations that GNT-MOVE excels in ensuring both cross-scene consistency and cross-view spatial smoothness. In Figure \ref{fig:cross-scene}, by aggregating expert selections from all test frames of the \{Trex, Horns\} scenes, we discern that experts 1\&3 are predominantly chosen for bright patterns, whereas experts 0\&2 are favored for darker or dimmer regions. Concurrently, Figure \ref{fig:cross-view} underscores that expert selections across varied views of the same scene exhibit layer-wise similarity and consistency.

\begin{figure}[h]
\begin{center}
\includegraphics[width=0.8\linewidth]{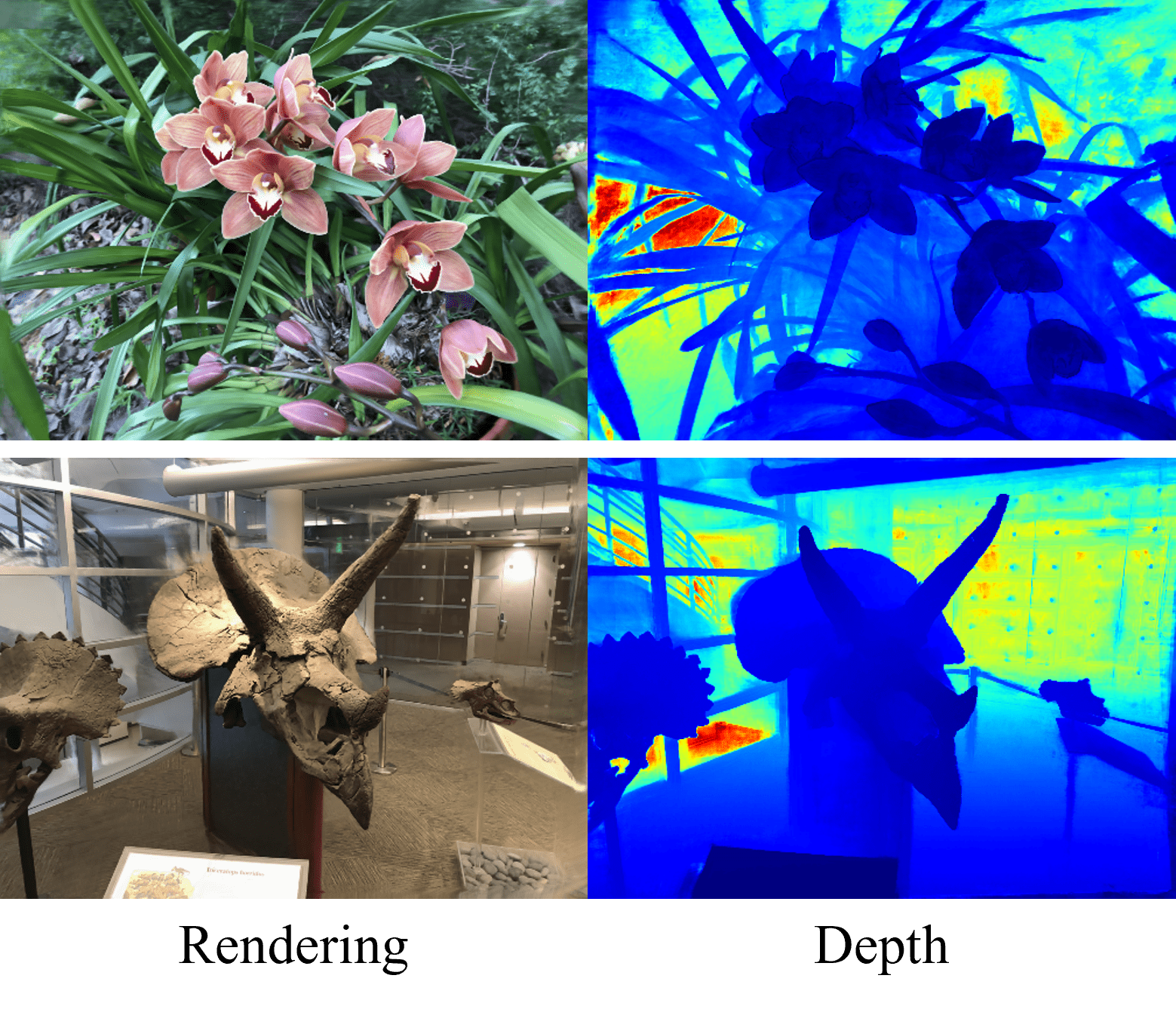}
\end{center}
\vspace{-4mm}
\caption{Geometry visualization. We show the depth maps from GNT-MOVE. Red indicates far and blue is near.}
\vspace{-2mm}
\label{fig:depth}
\end{figure}

Besides, following \cite{gnt}, we plot the depth maps computed from the learned attention values in Figure \ref{fig:depth}. The depth maps show clear physical ground that GNT-MOVE learns the correct geometry without explicit supervision. It also confirms that our geometry-aware smoothness does not distort or oversmooth the geometry.

\subsection{Ablation Studies} \label{sec:ablation}
We conduct ablation analysis on our key proposals, permanent expert and smoothness regularizer, on cross-scene generalization under zero-shot setting, and report results on the LLFF~\cite{mildenhall2019local} and NeRF Synthetic dataset~\cite{NERF} in Table~\ref{tab:ablation}.

As observed, directly plugging MoE into GNT cannot guarantee a good performance. We witness a performance drop on both datasets after adding the MoE. This is because MoE does not meet NeRF's cross-view consistency requirements and also does not learn the commodity across different scenes. Evidently, our customized design of permanent expert and smoothness regularizer both aid in improving model generalization capability. On the LLFF dataset, the smoothness regularizer brings the biggest performance gain, as cross-view consistency naturally benefits scenes with slightly disturbed views. On the NeRF Synthetic dataset with diverse complex scenes and materials, the permanent expert brings a considerable improvement as it enforces the commodity across scenes, thus contributing to the cross-scene consistency. 
Qualitative results in Figure \ref{fig:vis_expert} also illustrate their gains over the naive plug-in of MoE.

\vspace{-0.1em}
\section{Conclusion}
In this work, we focus on generalizable novel view synthesis on complex scenes and propose a novel learning-based framework, GNT-MOVE, that significantly pushes the frontier of this problem by introducing MoE to the domain of NeRFs. In order to better tailor MoE for generalizable NeRFs, we introduce a shared permanent expert and a spatial consistency objective to enforce cross-scene consistency and geometry-aware smoothness. GNT-MOVE proves its effectiveness by achieving SOTA performance on cross-scene generalization in both zero-shot and few-shot settings, on a broad collection of datasets. Our limitation is that we primarily focus on the view transformer of GNT, while introducing MoE into the ray transformer may be further promising - we regard it as future work.

{\small
\bibliographystyle{ieee_fullname}
\bibliography{egbib}
}

\pagebreak
\clearpage
\setcounter{equation}{0}
\setcounter{section}{0}
\setcounter{figure}{0}
\setcounter{table}{0}
\makeatletter
\renewcommand{\thesection}{S\arabic{section}}
\renewcommand{\theequation}{S\arabic{equation}}
\renewcommand{\thefigure}{S\arabic{figure}}
\renewcommand{\thetable}{S\arabic{table}}

\section{More Training / Inference Details}\label{train}

The base learning rates for the feature extraction network and GNT-MOVE are $10^{-3}$ and $5\times10^{-4}$, respectively, which decay exponentially over training steps. For the zero-shot generalization experiments, we train the network for 330,000 steps with 4096 rays sampled from 4 different views in each iteration. In the few-shot setting, we further fine-tune the pretrained model on each scene for 2,4000 steps. During the inference, we sample 192 coarse points per ray in all experiments.

\section{Cross-Scene Generalization}

\subsection{Testing Datasets} Local Light Field Fusion (LLFF)~\cite{mildenhall2019local} consists of 8 forward-facing captures of real-world scenes using a smartphone. NeRF Synthetic dataset~\cite{NERF} consists of 8, $360^{\circ}$ scenes of objects with complicated geometry and realistic material. Each scene consists of images rendered from viewpoints randomly sampled on a hemisphere around the object. Shiny-6 dataset~\cite{wizadwongsa2021nex} contains 8 forward-facing scenes with challenging view-dependent optical effects, such as the rainbow reflections on a CD, and the refraction through liquid bottles. Tanks-and-Temples~\cite{riegler2020free} is a complex outdoor dataset and contains large unbounded scenes. Following NeRF++, we evaluate on four scenes, including M60, Train, Truck, and Playground, and use the same evaluate views as NeRF++ does.
NMR~\cite{kato2018neural} contains 360\textdegree~views of various objects from unseen categories, which could be downloaded from {NMR\_Dataset.zip\footnote{https://s3.eu-central-1.amazonaws.com/avg-projects/differentiable\_volumetric\_rendering/data/NMR\_Dataset.zip} (hosted by the authors of Differentiable Volumetric Rendering \cite{niemeyer2020differentiable}). In the main paper, we report the average metrics across all eight scenes on each dataset for cross-scene generalization experiments.

\subsection{Per-Scene Breakdown Results for Zero-Shot Generalization}\label{breakdown}

To better demonstrate the effectiveness of our customized MoE, in Table \ref{tab:llff_breakdown} and Table \ref{tab:blender_breakdown}, we pick a few representative scenes for breakdown analysis of both GNT's and GNT-MOVE's quantitative results presented in Table 1a in the main paper.
The scenes we choose mainly cover the complex geometries (e.g., leaves and orchids) and materials (e.g., room and materials).
In both tables, our GNT-MOVE outperforms GNT by a significant margin in most scenes and achieves comparable results in the rest ones, demonstrating that with necessary customizations, MoE could be a strong tool to push the frontier of generalizable NeRF.

It is also worth mentioning that in Table \ref{tab:blender_breakdown}, our GNT-MOVE has demonstrated superior performance, especially in scenes with complex materials (\emph{e.g.}, Drums, Materials, Ship), showing that the customized MoE further enables cross-scene NeRF to generalize to difficult scenarios. 

\subsection{More Expert Selection Analyses}\label{expert}

In Figure \ref{fig:supp_exp}, we visualize more unseen scene rendering results and also the corresponding expert selections in the format of expert maps. It can be observed that our customized MoE is not only capable of keeping consistent selection across scenes (\emph{e.g.}, white background in the left three scenes, leaves in the right two scenes), but also reacts properly to complex lighting effects and materials (\emph{e.g.}, sparkling water in the left bottom scene Ship). 

\begin{figure*}[htb]
    \begin{centering}
        \includegraphics[width=0.9\linewidth]{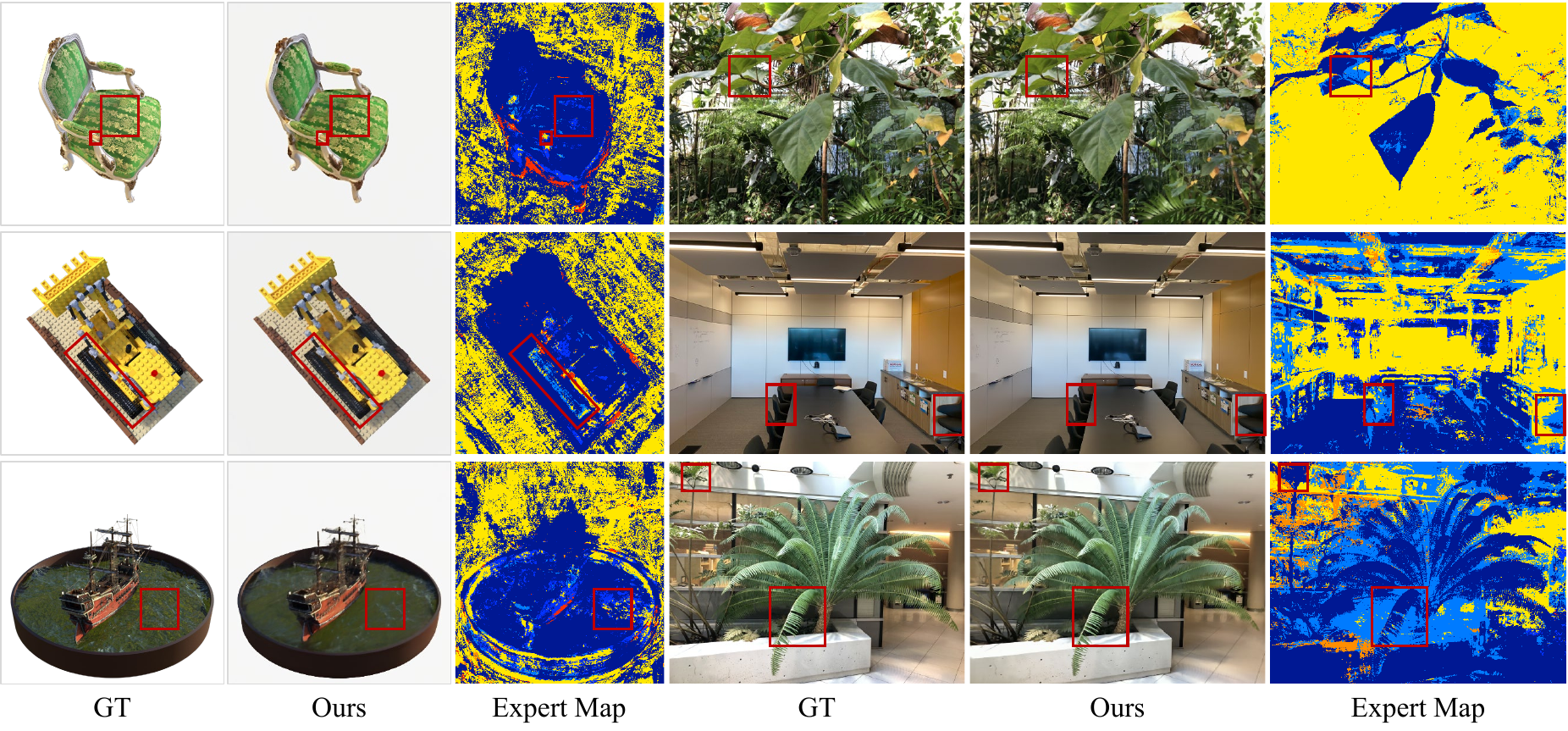}
        \par
    \end{centering}
    \caption{Results of unseen scene rendering and visualization of expert selection using different colors. }    
    \label{fig:supp_exp}
    \vspace{-2mm}
\end{figure*}

\begin{table*}[htbp]
    \centering
    \resizebox{\textwidth}{!}{
        \begin{tabular}{lcccc|cccc|ccc}
            \toprule[1.2pt]
            \multirow{2}{*}{Models} & \multicolumn{4}{c|}{Local Light Field Fusion (LLFF)}  &\multicolumn{4}{c|}{NeRF Synthetic}&\multicolumn{3}{c}{Tanks-and-Temples (Truck)}\\
            \cmidrule(r){2-12}
            & PSNR$\uparrow$ & SSIM$\uparrow$ & LPIPS$\downarrow$ & Avg$\downarrow$ & PSNR$\uparrow$ & SSIM$\uparrow$ & LPIPS$\downarrow$ & Avg$\downarrow$& PSNR$\uparrow$ & SSIM$\uparrow$ & LPIPS$\downarrow$\\
            \midrule[0.8pt]
            GNT & 25.86 & 0.867 & 0.116 & 0.047 & 27.29 & 0.937 & 0.056 & 0.030 & 17.39 & 0.561 & 0.429\\
            GNT-MOVE (E=1, K=4, w. PE) & 25.94&0.868&0.111&0.043&27.43 &0.939&0.057&0.029 & 19.08 & 0.611 & 0.393\\ \midrule
            GNT (Large) & 25.89&0.867&0.113&0.046&27.37&0.936&0.058&0.033 & 18.26 & 0.579 & 0.405\\
            GNT-MOVE (E=2, K=4, w. PE) & \textbf{26.02} & \textbf{0.869} & \textbf{0.108} & \textbf{0.043} &\textbf{27.47}  & \textbf{0.940} & \textbf{0.056}& \textbf{0.029}&\textbf{19.71}&\textbf{0.628}&\textbf{0.379}\\
            GNT-MOVE \textbf{w.o. PE} (E=3, K=5) & 25.81 & 0.867 & 0.114 & 0.047  &27.32  & 0.933 & 0.059 & 0.031 & 18.11 & 0.570 & 0.414\\
            \bottomrule[1.2pt]
        \end{tabular}
    } %
    \vspace{-1.5mm}
   \caption{Comparisons to illustrate the solid gain of MoE and PE in GNT-MOVE.}
    \label{tab:large}
\end{table*}

\begin{table}[htbp]
\centering
  \begin{subtable}[t]{0.8\linewidth}
  \centering
  \resizebox{\columnwidth}{!}{
  \begin{tabular}{lcccccc}
    \toprule
    Models & Room & Leaves & Orchids & Flower & T-Rex & Horns \\
    \midrule
    GNT & 29.63 & 19.98 &   18.84 & 25.86 & 24.56 & 26.34\\
    Ours & \textbf{29.94} & \textbf{20.45}  & \textbf{19.38} & \textbf{27.04} & \textbf{24.58} & \textbf{26.87}\\ 
    \bottomrule
  \end{tabular}}
  \caption{PSNR$\uparrow$}
  \end{subtable}
  \begin{subtable}[t]{0.8\linewidth}
  \centering
  \resizebox{\columnwidth}{!}{
  \begin{tabular}{lcccccc}
    \toprule
    Models & Room & Leaves & Orchids & Flower & T-Rex & Horns \\
    \midrule
    GNT & 0.940 & 0.756 & 0.661 & 0.859 & 0.885 & 0.892 \\
    Ours & \textbf{0.946} & \textbf{0.770}& \textbf{0.668} & \textbf{0.871} & \textbf{0.878} & \textbf{0.894}\\ 
    \bottomrule
  \end{tabular}}
  \caption{SSIM$\uparrow$}
  \end{subtable}
  \begin{subtable}[t]{0.8\linewidth}
  \centering
  \resizebox{\columnwidth}{!}{
  \begin{tabular}{lcccccc}
    \toprule
    Models & Room  & Leaves  & Orchids & Flower & T-Rex & Horns \\
    \midrule
    GNT & 0.091 & 0.183 & 0.216 & 0.108 & 0.127 & 0.118 \\
    Ours & \textbf{0.087}  & \textbf{0.173 }& \textbf{0.209 }& \textbf{0.101} & \textbf{0.123} & \textbf{0.113}\\ 
    \bottomrule
  \end{tabular}}
  \caption{LPIPS$\downarrow$}
  \end{subtable}
  \begin{subtable}[t]{0.8\linewidth}
  \centering
  \resizebox{\columnwidth}{!}{
  \begin{tabular}{lcccccc}
    \toprule
    Models & Room& Leaves & Orchids & Flower & T-Rex & Horns \\
    \midrule
    GNT & 0.031 & 0.097 & 0.119 & 0.048 & 0.054 & 0.046 \\
    Ours & \textbf{0.029}  & \textbf{0.093} & \textbf{0.115 }& \textbf{0.043} &\textbf{ 0.054} &\textbf{ 0.044}\\ 
    \bottomrule
  \end{tabular}}
  \caption{Avg$\downarrow$}
  \end{subtable}
  \caption{Comparison between our GNT-MOVE and GNT for cross-scene generalization under zero-shot setting on the LLFF Dataset (scene-wise).}
  \label{tab:llff_breakdown}
\end{table}

\begin{table}[htbp]
\centering
  \begin{subtable}[t]{0.7\linewidth}
  \centering
  \resizebox{\linewidth}{!}{
  \begin{tabular}{lccccc}
    \toprule
    Models & Chair & Drums & Materials & Mic & Ship \\
    \midrule
    GNT &  29.17 & 22.83 & 23.80 & 29.61 & 25.99\\
    Ours & \textbf{29.64} & \textbf{23.19}& \textbf{24.16} & \textbf{30.30} & \textbf{26.48}\\ 
    \bottomrule
  \end{tabular}}
  \caption{PSNR$\uparrow$}
  \end{subtable}
  \begin{subtable}[t]{0.7\linewidth}
  \centering
  \resizebox{\columnwidth}{!}{
  \begin{tabular}{lccccc}
    \toprule
    Models & Chair & Drums  & Materials & Mic & Ship \\
    \midrule
    GNT & 0.959 & 0.927 & 0.931 & 0.977 & 0.836 \\
    Ours  & \textbf{0.962} & \textbf{0.979 }& \textbf{0.935 }& \textbf{0.982 }&\textbf{ 0.845}\\ 
    \bottomrule
  \end{tabular}}
  \caption{SSIM$\uparrow$}
  \end{subtable}
  \begin{subtable}[t]{0.7\linewidth}
  \centering
  \resizebox{\columnwidth}{!}{
  \begin{tabular}{lccccc}
    \toprule
    Models& Chair & Drums  & Materials & Mic & Ship \\
    \midrule
    GNT& 0.038 & 0.059 & 0.058 & 0.017 & 0.154 \\
    Ours  & \textbf{0.038} & \textbf{0.057} & \textbf{0.056 }&\textbf{ 0.015 }& \textbf{0.149}\\ 
    \bottomrule
  \end{tabular}}
  \caption{LPIPS$\downarrow$}
  \end{subtable}
  \begin{subtable}[t]{0.7\linewidth}
  \centering
  \resizebox{\columnwidth}{!}{
  \begin{tabular}{lccccc}
    \toprule
    Models  & Chair & Drums  & Materials & Mic & Ship \\
    \midrule
    GNT & 0.021 & 0.044 & 0.040 & 0.014 & 0.054 \\
    Ours   & \textbf{0.021} &\textbf{ 0.042 }& \textbf{0.040 }& \textbf{0.013} & \textbf{0.051}\\ 
    \bottomrule
  \end{tabular}}
  \caption{Avg$\downarrow$}
  \end{subtable}
  
  \caption{Comparison between our GNT-MOVE and GNT for cross-scene generalization under zero-shot setting on the NeRF Synthetic Dataset (scene-wise).}
  \label{tab:blender_breakdown}
\end{table}

\section{More Comparison: GNT v.s. GNT-MOVE}

While the model size/speed is indeed not the main focus in this paper, GNT-MoE does generalize better, than the non-MoE counterpart with even heavier parameterization, while keeping per-instance inference low-cost.

Below, $\rhd$ \noindent \textbf{1)} and $\rhd$ \noindent \textbf{2)} demonstrate that the solid gain of MoE for generalizable NeRF goes way beyond naively larger model size; and $\rhd$ \noindent \textbf{3)} demonstrates that the gain can only be unleashed with PE and SR. Detailed results could be found in Table \ref{tab:large}. As preliminary, every expert in GNT-MOVE is half the size of GNT's same layer. The default GNT-MOVE (row 4) selects E = 2 such experts from K = 4 candidates, plus 1 permanent expert. Hence, if we treat the \underline{total parameter} and \underline{inference FLOPs} of GNT both as unit (\textbf{``1"}), then the default GNT-MOVE has \textbf{``2.5"} total parameter and \textbf{``1.5"} inference FLOPs. We construct the following comparison groups: 

$\rhd$ \noindent \textbf{1) the same FLOPs at inference.} Rows 1-2 compare GNT (FLOPs \textbf{``1"}) versus GNT-MOVE using only one selectable expert (E = 1) + one PE (0.5 + 0.5 = \textbf{``1"}). Despite the same inference complexity, the extra flexibility to ``select" endows GNT-MOVE with superior performance. 

$\rhd$ \noindent \textbf{2) the same total parameter.} Row 3 widens GNT by 2.5 times to match the total parameter size \textbf{``2.5"} of GNT-MOVE, called ``GNT (Large)". Compared to Row 4 (default GNT-MOVE), they have the same total parameter counts; meanwhile, GNT-MOVE has smaller per-inference FLOPs. However, GNT (Large) performs worse - and that clearly indicates for generalizable NeRF, ``the more parameter the better" is NOT the right quote, and per-scene specialization is necessary.

$\rhd$ \noindent \textbf{3) Does PE undermine MoE claim? NO.} First, the above two points already justified the necessity of MoE and disapprove ``natural to have better performance with more parameters". Second, our core claim is NEVER ``MoE shall work out of box for NeRF". Instead, while MoE is promising to balance ``generality” and ``specialization”, making it work for generalizable NeRF demands customized tactics to inject the key priors of cross-view consistency \& cross-scene commodity - PE is one such tactic.

To explain the second note, we stress that learning MoEs over NeRFs differs greatly from over standard image sets. If treating each view observation as an image sample, a ``NeRF dataset" would exhibit significant clustering due to different views of the same scene, and even different scenes will bear natural scene similarity. The highly non-i.i.d distribution, with multi-dimensional similarity entangled across views and scenes, can make naive MoE training more prone to collapse - see our ablation in Supplement sec. 3. Our important contribution is to show one can reap the benefit of MoE with proper regularizations including PE. 

To directly show PE values beyond just ``more parameters", we compare Row 5 in Table (replacing GNT-MOVE's PE with a selectable expert, and making E = 3), which has same total parameter \& inference FLOPs with our default GNT-MOVE setting (Row 4). Having PE helps evidently. 

\end{document}